\documentclass[11pt]{article}
\usepackage[margin=1in]{geometry}
\usepackage[T1]{fontenc}
\usepackage{lmodern}
\usepackage{microtype}
\usepackage{amsmath,amssymb}
\usepackage{array}
\usepackage{booktabs}
\usepackage{longtable}
\usepackage{graphicx}
\usepackage{placeins}
\usepackage{xcolor}
\usepackage{enumitem}
\usepackage[hidelinks]{hyperref}
\usepackage[capitalise,noabbrev]{cleveref}
\usepackage{natbib}
\graphicspath{{figures/}}
\setlist{nosep,leftmargin=1.4em}
\makeatletter
\newcounter{algorithm}
\def\fps@algorithm{tbp}
\def\ftype@algorithm{4}
\def\ext@algorithm{loa}
\def\fnum@algorithm{Algorithm~\thealgorithm}
\newenvironment{algorithm}{\@float{algorithm}}{\end@float}
\makeatother
\crefname{algorithm}{Algorithm}{Algorithms}

\newcommand{\ladder}{Ladder}          % records: H-gate
\newcommand{\mix}{Mixture}            % records: H
\newcommand{\mixpj}{Mixture-PJ}       % records: H-pj
\newcommand{\ruleopen}{Rule+Open}     % records: S
\newcommand{\openlvl}{Open}           % records: OPEN
\newcommand{\knnmem}{kNN-mem}         % records: B2
\newcommand{\napmem}{NAPPING-mem}     % records: B3
\newcommand{\unc}{+ask}               % records: +unc

\newcommand{\RQ}[1]{\textbf{RQ#1}}

\title{CRAFT: A Robot-Agent Harness\\ with Scope-Aware Adaptation}
\author{%
  \normalsize Peng Zhang, Jiawen Wang, Cheng Xue and Jochen Renz\\[0.4em]
  \normalsize Australian National University\\[0.4em]
  \small\texttt{\{peng.zhang, jiawen.wang, cheng.xue, jochen.renz\}@anu.edu.au}%
}
\date{}

\begin{document}
\maketitle

% Begin sections/00_abstract.tex
\begin{abstract}
A robot may know how to move an object yet still need to learn where it belongs.
When a person rejects a placement, the robot must decide whether the correction applies only to
that situation or expresses a broader rule. We study this problem in CRAFT (Contextual Requirement
Adaptation with Frozen Tools), a framework for learning local requirements without retraining robot
skills. It records physical success separately from whether an action meets the requirement.
Our \emph{scope ladder} uses memory when an action has already been judged in a nearly identical
situation. Elsewhere, it combines memory, explicit rules and a learner for requirements beyond those
rules, weighting their predictions by how well they predicted earlier feedback.
Controlled studies show gains over correction memory alone in utility, which rewards acceptable
actions and charges for mistakes and questions. In camera-grounded robot simulations, the ladder
asks fewer questions than language-model agents with persistent memory, but some of those agents
take fewer wrong actions and succeed more often after requirements change. All execution failures
and timeouts remain in the evaluation. These results support combining specific memories with
broader rules, while leaving reliable adaptation to changing requirements unresolved. All evidence
comes from simulation with generated requirements and simulated feedback.
\end{abstract}
% src: C7_RESULT_v1.txt §1 (i) +0.074 [+0.042,+0.108], (ii) +0.052 [+0.027,+0.079], (iv) +0.051 [+0.023,+0.083];
% C6_RESULT_v1.txt §3a, C8_RESULT_v1.txt §3e (replication);
% continuous_ladder_20261110/RESULTS_v1.md and results_v1/summary.json.
% droc_comparator_20261110/results_v1/{summary,audit}.json.

% End sections/00_abstract.tex

% Begin sections/01_intro.tex
\section{Introduction}
\label{sec:intro}

\begin{figure}[t]
  \centering
  \includegraphics[width=\linewidth]{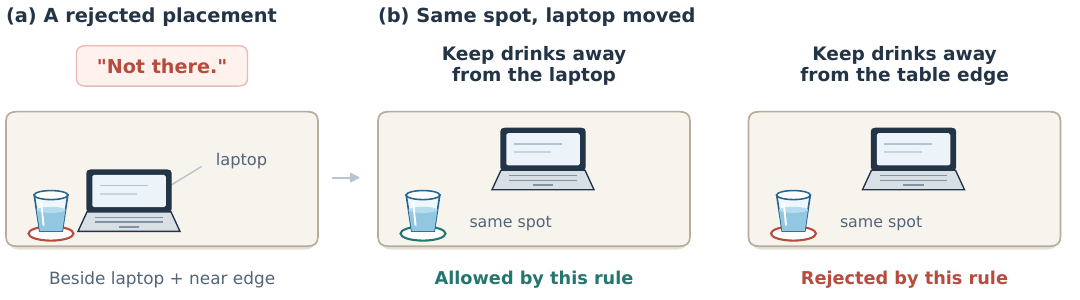}
  \caption{The reach of a correction: an illustrative household scenario. (a) A drink is placed beside
  a laptop, near the table's edge, and the person objects. (b) The two panels show the same arrangement
  after the laptop moves to the centre. The original spot satisfies one possible rule but violates
  the other. One correction does not distinguish the rules.}
  \label{fig:teaser}
\end{figure}

Consider a robot bringing a glass of water to someone working at a table. It sets the glass beside
a laptop, near the table's edge. The person says, ``Not there.'' Should drinks be kept away from
the laptop, or away from the edge? Both rules explain the objection. Once the laptop is moved to
the centre of the table, they disagree about whether the original spot is acceptable
(\cref{fig:teaser}).

We call this the correction's \emph{reach}: the situations to which it should apply. A correction
interpreted too narrowly allows the same mistake in a new scene; one interpreted too broadly rules
out acceptable actions. This problem arises when a robot has usable physical skills but must learn
unstated requirements at the site where it operates. Skill libraries and robot-agent harnesses make
it possible to reuse existing controllers \citep{hu2026halo}.

Memory and rules offer different ways to use that feedback. Memory can recall what happened in a
similar situation. A rule can extend the lesson to unfamiliar objects or layouts. Each has a weakness:
similar-looking scenes may require different actions, and a rule may be inferred too broadly or
applied incorrectly because perception is noisy.

We study this choice in CRAFT (Contextual Requirement Adaptation with Frozen Tools). CRAFT is a
robot-agent framework that keeps physical skills fixed while allowing the method that learns local
requirements to change. It separates whether a motion succeeded from whether the chosen action was
acceptable. A failed movement therefore supplies no verdict about the local requirement. Compared
methods receive the same kinds of observations, feedback and opportunities to ask questions; each
learns only from the outcomes of its own actions.

Within CRAFT, we propose the \emph{scope ladder}. If an action has already received feedback in a
nearly identical situation, memory decides. Otherwise, the ladder combines predictions from memory,
an explicit set of possible rules, and an additional learner for requirements those rules cannot
express. Each predictor is weighted by how well its past predictions matched the feedback that
followed. This gives specific experience priority where it applies and allows broader
rules to guide decisions elsewhere. The arbitration rule combines established memory, rule-learning
and weighting techniques.

Existing systems already learn reusable preferences from corrections or maintain revisable user
and scene memories \citep{zha2024droc,wang2026carryon}. Our focus is the choice between specific
memories and broader predictions when feedback consists of sparse yes/no verdicts. We evaluate
the method in controlled studies and in a continuing service that carries experience across robot jobs.

\paragraph{Contributions.}
\begin{itemize}
  \item \textbf{A formulation and shared framework} (\cref{sec:problem}). We formalise learning the
    reach of action feedback and provide an engineering framework that keeps physical execution,
    task acceptability and the cost of asking separate.
  \item \textbf{A method for combining memory and rules} (\cref{sec:method}). The scope ladder gives
    memory priority on near-identical situations and combines predictions elsewhere according to
    their earlier predictive performance.
  \item \textbf{Evidence of benefits and limits} (\cref{sec:results,sec:integrated}). Pre-registered
    studies find higher utility than correction memory alone on new and repeated situations.
    Component comparisons link generalisation to rule inference and repeated-situation performance
    to the memory gate. Descriptive robot-simulation comparisons with a frontier agent and
    adaptations of DROC and CarryOn find lower query costs for the ladder, alongside tradeoffs in
    wrong actions and completion after requirements change. All failures remain in the evaluation.
\end{itemize}

All evidence comes from simulation with generated requirements and simulated feedback. The robot
comparisons use a shared frozen skill, camera-derived observations and external scene setup;
question counts do not measure human effort. They test adaptation within this setting, and reliable
revision after unannounced requirement changes remains unresolved. Unsuccessful studies are
retained in the appendices.

% End sections/01_intro.tex

% Begin sections/02_related.tex
\section{Related work}
\label{sec:related}
% Full-text audits: RELATED_WORK_FULLTEXT_CHECK_v1.md and LITERATURE_UPDATE_v1.md.

\paragraph{Harnesses around frozen policies.} Persistent memory and revision around frozen robot policies
are established directions. CarryOn \citep{wang2026carryon} is particularly close at the system level:
it combines revisable user and scene memories with a frozen VLA, proactive shielding, recovery skills
and reflection. Lasting preferences are extracted from dialogue; explicit revisions replace or remove
profile entries, and observations trigger scene-memory updates. CRAFT builds on this use of persistent, revisable memory. We ask how sparse action verdicts
should generalise when they leave the intended requirement and its scope unstated.
Our paired comparison evaluates the ladder against an adaptation that retains CarryOn's native memory stores,
write/delete operations and context policies under our shared verdict-feedback interface
(\cref{app:carryon}); it does not evaluate CarryOn's shielding and recovery.

Harness VLA \citep{zhang2026harnessvla} turns a frozen VLA into a retryable
contact-rich primitive alongside analytic primitives, and keeps task-specific traces and global success
rules and failure models extracted from its own exploration. The RoboHarness of
\citet{li2026roboharness} retrieves
success and failure precedents by CLIP similarity for a VLM orchestrator that diagnoses failures and selects
interventions. A distinct RoboHarness, by \citet{huang2026roboharness}, orchestrates heterogeneous policies
using capability descriptions and execution memory; its Memory Bridge addresses incompatible states
at policy handoffs without joint retraining. RoboArchitect \citep{wu2026roboarchitect} has a coding model
revise the harness around a frozen policy from rollout traces, retaining lessons for later executions.
These works address overlapping deployment problems through different feedback and control interfaces.
CRAFT uses these established components through contracts that let adaptation methods be swapped
under equal information, with execution and acceptability reported separately.
HALO \citep{hu2026halo} illustrates the skill layer such a harness can use. It learns whole-body humanoid
skills (walking to named objects, sitting, crawling, pushing boxes) solely from data synthesised in
procedurally generated homes, then transfers them zero-shot to unseen rooms. Its VLM orchestrator chains
skills from a spoken request, retries failed steps and may ask the operator. The reviewed version plans
from the current instruction and camera frame, describes its orchestration as relatively simple, and
reports no memory of a person's statements across requests. HALO provides the kind of frozen skills
CRAFT assumes; learning a correction's reach is a separate problem.
% src: arXiv 2611.06908v1 (HTML read 10 Nov 2026): App. D3-D4 (plan from instruction + egocentric frame;
% actions incl. retry_step, ask_operator); Sec. V-1 "our orchestration layer is relatively simple".

\paragraph{Learning preferences and rules from few examples.} TidyBot \citep{wu2023tidybot} summarises a
person's example placements into general rules with an LLM and applies them to unseen objects. SYNAPSE
\citep{modak2025synapse} synthesises preferences as programs in a neuro-symbolic language over images, from
demonstrations paired with explanations, and generalises well out of distribution; it does not handle
preferences that change over time. Hypothesis Search \citep{wang2024hypothesis} has an LLM propose
natural-language hypotheses, implement them as Python and keep those that reproduce the given examples;
hypothesis refinement \citep{qiu2024phenomenal} finds LMs better at proposing rules than at applying them.
NeatNet \citep{kapelyukh2021neatnet} learns a latent preference vector from many users' arrangements. These
methods learn from demonstrations or example sets; our setting supplies online yes/no verdicts for physically completed actions,
with budgeted full-menu labels during commissioning and optional queries. Our rule level follows
Bayesian concept learning over a grammar \citep{goodman2008rational}. Comparators include an LLM rule
writer inspired by Hypothesis Search (an 8B model and, on six streams, a frontier model), a VLM
in-context learner and Popper \citep{cropper2021popper}. A separate integrated pilot compares against a frontier
agent combining retrieved instances, editable rules and its own execution history (\cref{sec:integrated}).

\paragraph{How far a correction reaches.} DROC \citep{zha2024droc} is a close correction-learning
comparator: an LLM distils language corrections into plan- and skill-level knowledge. Its plan-level
implementation retrieves relevant constraints semantically; task-category and visual retrieval also
support object grounding and skill reuse. It explicitly studies generalisation of preferences and
constraints, which overlaps with CRAFT's learning objective. CRAFT tests
prequential arbitration between predictors under sparse verdict feedback. DROC's described modules do
not use that arbitration rule. ExpeL \citep{zhao2024expel} likewise uses both its extracted insights and its recalled
trajectories, in text domains. Recurrent Summary Memory (RSM) \citep{wang2026rsm} compresses interaction history
into a fixed-size recurrent summary for in-context reinforcement learning; CRAFT studies when feedback
from past actions should apply to new situations.
\citet{lee2026flexible} address reach directly: a language model edits a
robot's task features after each language correction, with an explicit operation that generalises a
specific feature to the correction's intent. In their work an LM decides how far a correction reaches; in
ours each level's weight depends on how well it predicted earlier verdicts
\citep{freund1997specialize,herbster1998tracking}, and we measure prospectively when each level should decide.
Earlier, \citet{bajcsy2018onefeature} inferred which feature a physical correction targets and
\citet{bobu2018misspecified} acted conservatively when the hypothesis space could not explain a correction;
our open level plays that role for rules outside the grammar.

\paragraph{Asking for help.} Uncertainty-triggered queries are standard in active learning
\citep{settles2009active} and in planners that ask when unsure \citep{ren2023knowno}. APRICOT
\citep{wang2024apricot} keeps LLM-proposed candidate preferences from demonstrations, updates a Bayesian belief
from yes/no answers and chooses questions by information gain. We use an uncertainty band for asking.
We did not compare it with APRICOT; a tested scope-aware question selector did not improve performance
(\cref{app:negative}).

\paragraph{Safety constraints from feedback.} A VLM can turn language feedback and images into constraints for
an online reachability filter \citep{santos2025updating}; sparse ``unsafe'' flags can define a
nearest-neighbour region of likely failures via conformal prediction \citep{feldman2025conformal}; and an
LLM can ground fixed safety rules into temporal-logic specifications that guard a plan
\citep{ravichandran2026roboguard}. We tested the reach question under asymmetric harm once; that study failed
its validity check (\cref{app:negative}), and we make no safety claim.

\paragraph{Open-world adaptation and evaluation.} NovPhy \citep{pinto2024novphy} evaluates how agents
detect and adapt to novel physical conditions. \citet{pinto2022measuring} argue that open-world
evaluation should measure both the timeliness of novelty detection and the quality of adaptation.
CRAFT examines how sparse verdict feedback helps identify acceptable actions as requirements change
around fixed robot skills. NAPPING \citep{xue2023napping} adapts a trained reinforcement-learning
policy only where it stops working, by learning adaptation principles for those regions; the harness's correction
memory (\napmem) and its staleness rule are derived from it.
Freezing an analysis plan before data exist \citep{nosek2018preregistration} is uncommon in robot learning. We adopt it for every confirmatory study and report all outcomes.

% End sections/02_related.tex

% Begin sections/03_problem.tex
\section{Problem and setting}
\label{sec:problem}
% src: RESULTS_NARRATIVE_v1.md §1-2; C4-C7 protocols; ARCHITECTURE_v1.txt (4 Nov 2026); K1 v2 / K2 v1 contracts.

\subsection{Learning a correction's reach}
Returning to the household example in \cref{fig:teaser}, suppose the robot can place
a glass of water beside a laptop near the table's edge, or in a clear space in the
middle. The household's rule is to keep drinks away from the laptop, but this rule
has not been stated to the robot. It places the glass beside the laptop, and the
person says, ``Not there.'' The placement succeeds physically but violates the
requirement. The robot receives no explanation and no verdict about the unchosen
location. Both ``keep drinks away from the laptop'' and ``keep drinks away from the
table's edge'' fit this feedback.

On the next delivery, the laptop has moved to the middle of the table. The original
spot is now away from the laptop but still near the edge. Under the first explanation,
that spot is acceptable; under the second, it remains unacceptable. The household
rule has not changed, only the layout. The robot must decide whether the earlier
rejection still applies. This is the question of a correction's \emph{reach}.

Each delivery is a job, the observed table arrangement is its context, and the
candidate placements form the action menu. We formalise these repeated decisions
as follows.

A \emph{stream} is a sequence of $T$ jobs at one site. At job $t$ the robot observes a context $x_t$ (an
RGB-D image of the scene) and a menu $\mathcal{A}_t$ of admissible actions. The actions differ in one
\emph{role}: which object to move, or which target to put it on. Actions are drawn from the declared skill menu, but execution can fail; the requirement determines
whether an action is \emph{acceptable} under a site requirement that nobody stated. A hidden rule $r$
maps each context to its acceptable set $r(x_t)\subseteq\mathcal{A}_t$, which may be empty.

On each job the provider either executes one action $a_t\in\mathcal{A}_t$ or holds off. Under the feedback
level used throughout (F1), this defines a contextual partial-monitoring problem: the provider observes the
bandit label $y_t=[a_t\in r(x_t)]$ only for the single action it executed, while holding off and failed
executions return no verdict, and nothing is revealed about unexecuted actions $a\neq a_t$. In the decision-level studies, $T=40$: the first four
jobs supply full-menu commissioning labels and the remaining 36 are scored with net utility
\[
  u_t = \begin{cases} +1 & \text{execution completes and } a_t \in r(x_t),\\ -1 & \text{execution completes and } a_t\notin r(x_t),\\ \phantom{+}0 & \text{held off or execution failed,}\end{cases}
\]
a decision-quality score rather than physical throughput. A provider with the right to ask may instead spend
one of three questions per stream, at a cost of $0.5$ each, to receive full-information labels
$\{y_t(a)\}_{a\in\mathcal{A}_t}$ for every action in the current context.
The continuing-service pilot uses $T=24$, reports accepted completion over all 20 scored assignments,
and accounts separately for physical failures, wrong dispatches and intervention costs (\cref{sec:protocol}).

A verdict on $(x_t, a_t)$ is informative only through its \emph{reach}: the set of later context-action pairs
$(x_{t'},a)$ whose predicted acceptability $p(a\mid x_{t'})$ it changes. Every learner fixes a reach
implicitly. Nearest-neighbour memory extends a verdict to a local ball around $(x_t,a_t)$ in visual feature
space; a rule learner extends it across every pair $(x_{t'},a)$ that shares the predicates of its surviving
hypotheses. Because perceptual facts are noisy and rules may fall outside any fixed grammar, no single reach
dominates across a stream. The problem is to arbitrate online, from sparse bandit verdicts, which reach to
trust for each candidate action.

The decision-level providers use frozen features and perceived facts derived from RGB-D, public menu
names, an object catalogue and a declared grammar. The integrated providers share a camera-derived
fact frontend and frozen-feature memory; the frontier receives facts and retrieved records, not raw RGB. Simulator state,
evaluator predicates, hidden rule revisions and outcomes of unexecuted actions are never inputs.

\subsection{The CRAFT harness}
\begin{figure}[t]
  \centering
  \includegraphics[width=\linewidth]{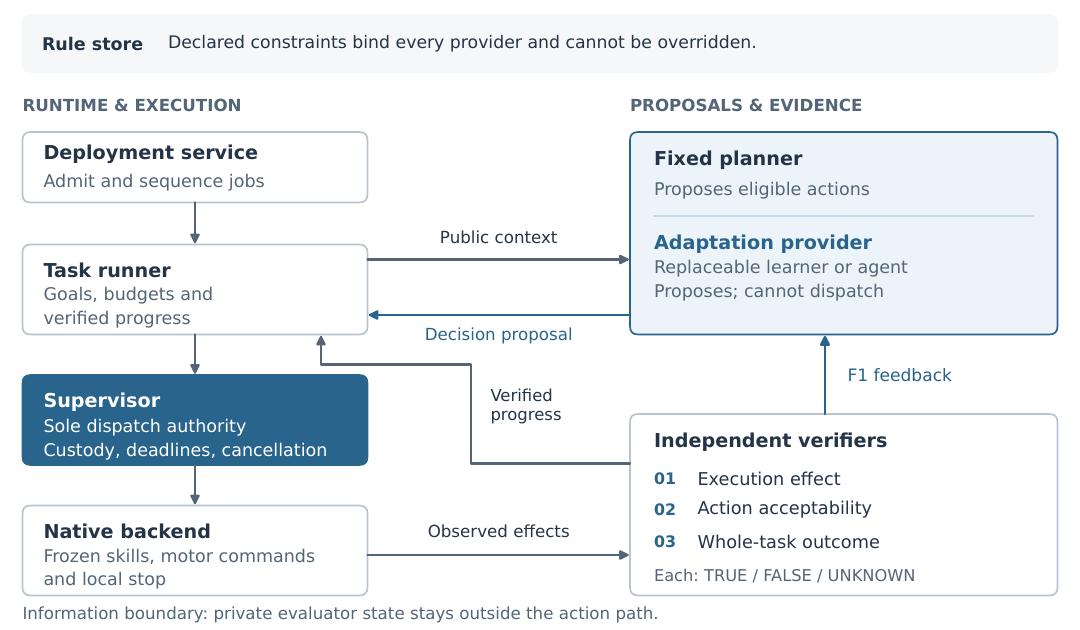}
  \caption{Authority in CRAFT. The adaptation provider sees only public evidence and proposes; only the
  supervisor dispatches; independent verifiers report the effect, the acceptability of the action and the
  whole-task outcome as separate verdicts; the provider receives the F1 feedback record for the action it
  executed. Private oracle state is excluded; labels reach the provider only through declared feedback channels.}
  % src: ARCHITECTURE_v1.txt §2; make_figures.py
  \label{fig:arch}
\end{figure}

CRAFT runs repeated, authorised jobs at one site with independently developed skills. It verifies progress,
retains experience and permits bounded adaptation, without making any memory or learning algorithm a runtime
dependency. Each component has a defined authority (\cref{fig:arch}). A deployment service admits jobs, and a task
runner owns each job's goals, verified progress and budgets. Planner and adaptation providers
\emph{propose} actions; a supervisor controls dispatch, physical custody, deadlines and cancellation.
Independent verifiers supply the evidence for progress. Declared site rules
live in a rule store and bind every provider equally: a learned proposal cannot relax a declared constraint,
and a physical success cannot remove a prohibition.

\paragraph{Contracts that make providers comparable.} In the architecture, a provider receives a
\texttt{DecisionContext} (fresh public evidence, site cues, goals, declared constraints, the qualified
actions and its own factual history), returns a \texttt{DecisionProposal} (execute, observe, wait, ask for
help or finish, with a reason and the cost it charges), and later receives a \texttt{FeedbackRecord} in which
the \emph{execution effect}, the \emph{acceptability} of the action and the \emph{whole-task} verdict are
separate fields, each TRUE, FALSE or UNKNOWN. A single failure bit would not tell a learner whether the
motor, the destination or the remembered rule was wrong; keeping them apart lets a requirement learner ignore
execution failures instead of learning false rules from them. The feedback contract fixes what a provider may
learn, here the F1 level above. Asking a person is a separate, charged channel.

\paragraph{Information boundary.} For the selector providers in the initial repeated-context, new-context and component studies, three tests ran before any
test output: a \emph{canary} test (private fields replaced by sentinel values must leave every decision
bit-identical), a \emph{static import} check (provider modules cannot import the label generator, rule
evaluator or bank loader) and a \emph{prefix-invariance} test (decisions up to job $t$ must not change when
everything after $t$ is randomised). Later providers had study-specific checks; we found no evidence of label
leakage, but coverage is not uniform (\cref{app:harness}).

\paragraph{What is implemented.} The supervisor, skill registry, native backends, task runner and deployment
service exist as bounded prototypes. The main confirmatory studies use a decision-level driver.
The initial closed-loop demonstration launched a fresh simulator for each job. The continuing-service pilot uses the same task runner, supervisor and
deployment service for both providers, retaining native state throughout each stream. \Cref{app:harness} gives the details.

% End sections/03_problem.tex

% Begin sections/04_method.tex
\section{The scope ladder}
\label{sec:method}
% src: RESULTS_NARRATIVE_v1.md §1-2 (corrected 8 Nov); C4/C5/C6/C7/C8 protocols; FRAMING_SECTION_v1.md.

For every menu action $a$, each of three predictors estimates the probability
$p_c(a\mid x_t)$ that $a$ is acceptable. The ladder determines how to combine these predictions.

\subsection{Perception and three predictors}
\paragraph{Perceived facts.} A probe trained on separate scenes reads, for each catalogue object, whether it
is present, its colour and its tabletop position from a frozen vision backbone \citep{shukor2025smolvla}; kind
and movability come from fixed name tables. We write $f_t$ for the resulting fact vector and $z_t$ for the
pooled, normalised visual/depth features using the same backbone. The integrated R1 pilot replaces
this probe with a shared segmentation/depth frontend and its own disjoint feature normalisation;
the learner equations are unchanged (\cref{app:continuing}).

\paragraph{Rule level $G$.} A hand-written grammar $\mathcal{H}$ enumerates rules over perceived facts in five
families: attribute (colour or kind), spatial (nearest, farthest, leftmost, rightmost), occupancy (no
movable object within radius $\rho$ of the target), presence (one rule if a cue object is present, another
otherwise) and composite (a primary rule with a distance threshold and a spatial fallback), about 1{,}800 rules
per site. The provider initialises weights $w_h$ to a family-balanced prior $\pi_0(h)$ (\cref{app:settings}), applies
fixed-share forgetting \citep{herbster1998tracking} at rate $\eta=0.002$ before each job after the
first, and updates on each observed verdict $(x,a,y)$ with a noisy-label likelihood:
\begin{equation}
  w_h \leftarrow (1-\eta)w_h + \eta\,\pi_0(h),\qquad
  w_h \leftarrow \frac{w_h\,\ell_h(y\mid x,a)}{\sum_{h'} w_{h'}\ell_{h'}(y\mid x,a)},\qquad
  \ell_h=\begin{cases}1-\varepsilon & \text{if } [a\in h(f)] = y,\\ \varepsilon & \text{otherwise,}\end{cases}
  \label{eq:posterior}
\end{equation}
with $\varepsilon=0.05$, and predicts $p_G(a\mid x_t)=\sum_h w_h\,[a\in h(f_t)]$. This is predictive rule
inference in the sense of \citet{goodman2008rational}; it does not identify causes.

\paragraph{Open level $O$.} Per action, an L2-regularised logistic regression on $f_t$. It can fit rules
outside the grammar, within the reach of a linear function of the facts.

\paragraph{Memory level $M$.} Per action, a $k$-nearest-neighbour vote ($k=3$) over the provider's own labelled
records $(z, a, y)$, weighted by cosine distance $d(z_t, z)$ on the pooled features.

\subsection{Evidence-weighted mixture}
The three predictions are mixed with weights proportional to a prior $\pi_c$ times the exponential of each
predictor's accumulated prequential evidence \citep{freund1997specialize}:
\begin{equation}
  p_{\mathrm{mix}}(a\mid x_t)=\sum_{c\in\{G,O,M\}} \omega_c\, p_c(a\mid x_t),\qquad
  \omega_c\propto \pi_c \exp\Big(\sum_{i\in\mathcal{L}_{<t}}\log \tilde p_c(y_i\mid x_i,a_i)\Big),
  \label{eq:mixture}
\end{equation}
with $\pi=(0.6,0.2,0.2)$ and $\mathcal{L}_{<t}$ indexing every earlier revealed action label, including
commissioning and queried alternatives. Each term $\tilde p_c(y_i\mid x_i,a_i)$ is the smoothed predictive
likelihood level $c$ assigned to verdict $y_i$ \emph{before} updating on it: for $G$, the marginal
noisy-label likelihood $\sum_h w_h \ell_h(y_i\mid x_i,a_i)$; for $O$,
$1-\varepsilon_{\mathrm{open}}=0.75$ when its hard prediction matches $y_i$ and $0.25$ otherwise
($0.5$ before any label has been seen for $a_i$); and for $M$, its pre-insertion vote probability for
$y_i$ clipped to $[0.05, 0.95]$ (\cref{app:settings}). Predictors assigning higher likelihood to incoming
verdicts therefore gain relative weight online; we make no calibration claim for $p_{\mathrm{mix}}$.

\subsection{The gate}
Global prequential weights $\omega_c$ pool prediction errors across the entire stream. When new and repeated
situations are interleaved, evidence from new contexts can give rule inference enough weight to overrule
memory on an exact repeat, even when the probe facts $f_t$ are wrong. The ladder therefore treats
near-identical memory as a specialist \citep{freund1997specialize} that takes precedence on its support:
\begin{equation}
  p(a\mid x_t)=\begin{cases} p_M(a\mid x_t) & \text{if } \exists\,(z,a,y)\in\mathcal{M}:\ d(z_t,z)<\tau,\\
  p_{\mathrm{mix}}(a\mid x_t) & \text{otherwise,}\end{cases}
  \label{eq:gate}
\end{equation}
with $\tau=0.01$. ``Near-identical'' is a fixed metric threshold, not a learned gate. A variant that instead keeps
separate evidence for seen and new pairs (\mixpj) was the primary arm of the component study; the gated ladder was promoted
to primary after development studies and tested in the main confirmatory study.

\paragraph{Decision and asking.} Every provider shares one decision layer: take $a^\star=\arg\max_a p(a\mid
x_t)$, and hold off if $p(a^\star\mid x_t)<0.5$. A provider with the right to ask uses an uncertainty band
\citep{settles2009active,ren2023knowno}: ask if $0.25<p(a^\star\mid x_t)<0.75$ and a question is left. The
same channel, budget, cost and rule apply to every arm that may ask. \Cref{alg:ladder} summarises one job
(a diagram is in \cref{app:settings}, \cref{fig:ladder}).

\begin{algorithm}[t]
\centering
\fbox{\begin{minipage}{0.95\linewidth}\small
\textbf{Input:} context $x_t$, menu $\mathcal{A}_t$, rule weights $w$, evidence totals $E_c$, memory
$\mathcal{M}$, questions left $q$.
\begin{enumerate}[label=\arabic*:,leftmargin=2em,itemsep=1pt]
  \item \textbf{if} $t>1$: $w_h \leftarrow (1-\eta)w_h + \eta\,\pi_0(h)$ for all $h\in\mathcal{H}$
    (fixed-share forgetting).
  \item $f_t,z_t \leftarrow$ probe and features of $x_t$.
  \item \textbf{for} $a\in\mathcal{A}_t$: compute $p_G,p_O,p_M$; set $p(a)$ by \cref{eq:gate}, using
    \cref{eq:mixture} with $\omega_c\propto\pi_c e^{E_c}$ when no record of $a$ lies closer than $\tau$.
  \item $a^\star\leftarrow\arg\max_a p(a)$.
  \item \textbf{if} $q>0$ \textbf{and} $0.25<p(a^\star)<0.75$: ask; receive $y(a)$ for every
    $a\in\mathcal{A}_t$; $q\leftarrow q-1$; act on an acceptable action if there is one.
  \item \textbf{else if} $p(a^\star)\ge0.5$: execute $a^\star$; receive $y(a^\star)$ unless execution failed.
  \item \textbf{else} hold off (no verdict).
  \item \textbf{for} each received verdict: add $\log\tilde p_c$ to $E_c$ (memory scored before storing);
    update $w$ by the Bayes step of \cref{eq:posterior}; refit the open level; store $(z_t,a,y)$ in $\mathcal{M}$.
\end{enumerate}
\end{minipage}}
\caption{The scope ladder on one scored job (with asking). Without the right to ask, step 5 is skipped.}
\label{alg:ladder}
\end{algorithm}

\subsection{Staleness rule}
The harness's correction memory, derived from NAPPING \citep{xue2023napping}, has a staleness rule: when an
unacceptable verdict arrives for action $a$, it removes earlier acceptable records of $a$ with the same
perceived-fact key. Added to the ladder's memory, it targets the case where a remembered ``yes'' was recorded
under a rule that has since changed; it did not fire in the evaluated fixed-rule patterns A and O. It is a separate variant, tested in
the staleness study without asking.

% End sections/04_method.tex

% Begin sections/05_setup.tex
\section{Experimental setup}
\label{sec:protocol}
% src: STUDY_INDEX.md §1; C6/C7/C8/C10 protocols and results; C7 protocol §2 (asking).

We ask five questions. \RQ{1}: does the ladder improve net utility over correction memory on fresh sites,
with and without the right to ask? \RQ{2}: do the gains replicate, and which components account for them?
\RQ{3}: how does it compare with other rule learners? \RQ{4}: what happens when the site's rule changes?
\RQ{5}: how does the ladder compare with model-based adaptation when both run in closed loop through
the continuing service?

\paragraph{Decision-level sites and scenes.} Each site is a rendered tabletop scene composition \citep{liu2023libero} with
50--75 contexts. Rules and acceptable sets come from a generator over the grammar's five families. On test
sites the perception probe reads presence with about 97\% accuracy, colour with about 99\% (70\% on assets
whose colour was swapped) and position with a median error of about 4\,cm. % src: C7_RESULT §header: 96.6%, 99.1%, 3.9 cm

\paragraph{Regimes and patterns.} In regime \emph{N} every job shows a new context; in regime \emph{R}, 20
contexts are drawn 40 times, so about 62\% of jobs repeat a context. Pattern \emph{A} (primary) is a fixed rule
from one of the generator's families; \emph{C} is as A, but the rule changes without announcement to another
rule of the same family at job 20; \emph{E} is as A with 20\% execution failures, which return no feedback; and
\emph{O} is a fixed rule from separately written forms that the full grammar cannot express on true facts (out
of grammar). In A, C and E the family that generated the stream's rule is removed from the provider's grammar,
including every rule that nests it, so the provider never holds the true rule; in O it has the full grammar.

\paragraph{Arms.} \emph{\knnmem}: kNN correction memory on the same frozen features, with $k$ and recency
weighting chosen on development sites per regime; it is the standard site-scoped retrieval baseline that any
method must beat. \emph{\napmem}: the harness's NAPPING-derived correction memory; its distance threshold was
computed from the test sites' perceived facts before the streams ran (no labels), a transductive exception
that favours this baseline. \emph{\openlvl}: the open level alone. \emph{\ruleopen}: the rule and open levels
without memory. \emph{\mix}: the evidence-weighted mixture of \cref{eq:mixture}, without the gate.
\emph{\mixpj}: the mixture with separate evidence per seen and new pair. \emph{ILP}: Popper
\citep{cropper2021popper} over a restricted predicate vocabulary built from the same perceived facts, in one
configuration chosen on tuning sites; its hypothesis class differs from the grammar's (for example, one
occupancy radius instead of three). \emph{VLM}: Qwen3-VL-8B \citep{qwen3vl} as an in-context correction learner
over earlier scenes and their verdicts. \emph{LLM rule writer}, inspired by Hypothesis Search
\citep{wang2024hypothesis}: an LLM writes candidate rules as Python functions, refines them after wrong
predictions and keeps the one most consistent with the verdicts, with the same 8B model and, on six streams, a
frontier model. \emph{Oracle} (the true rule) and \emph{\ruleopen@true} (rules on true rather than perceived
facts) are references. Suffix \emph{\unc} marks an arm with the right to ask. Settings are in
\cref{app:settings}.

\paragraph{Pre-registration.} A \emph{confirmatory} study tests a pre-specified claim. An author approves its protocol,
code and analysis plan before they are frozen and hashed (SHA-256), before any outcome data exist. Each online confirmatory study used 150 fresh sites built after the freeze with a new seed,
excluding near-duplicates of earlier sites, and evaluated once. Success or failure is determined by conditions set in
advance; deviations are logged before outputs are read, and a failed study is never reclassified as successful. A \emph{development}
study uses sites already used in development (100 in most, 30 in the later model comparisons); it
has a written plan with interpretation criteria, but supports no confirmatory claim. A study whose pre-set validity check fails is
reported as ``not a result''. \Cref{app:ledger} summarises the evidence; the complete
study ledger is retained in the project archive.

\paragraph{Decision-level statistics.} The unit is the site. We report the mean over sites of per-stream net utility over the
36 scored jobs, and 95\% bootstrap confidence intervals over sites with 2{,}000 resamples
\citep{efron1993bootstrap}. Contrasts are paired by site. Non-inferiority margins were set in each protocol.

\paragraph{Where the studies run.} Confirmatory and development studies run providers in a study driver at the
decision level: perception runs on rendered scenes, but actions are not executed physically, and feedback
comes from the generator. The initial closed-loop demonstration and the continuing-service comparisons execute native simulated actions. Their populations and outcomes
are reported separately; the pilot is not another replication of the 150-site studies.

\paragraph{Continuing-service frontier comparison.} Six paired streams each contain four commissioning and
20 scored jobs per method: 288 assignments, 120 scored per method. A fixed-pelvis R1 pushes one of
three or four visible items from a six-object catalogue. Layouts, object subsets and a white cue vary;
three streams keep their requirement and three change it without announcement. Each method retains
its robot, service and learner state within a stream; external object setup is logged every job.
The unchanged deployment service, task runner, supervisor, camera frontend and frozen push skill
serve both providers. A prospective skill screen passed 14/16 with both failures retained.

The comparator is \texttt{gpt-6.1-sol} at medium reasoning with retrieved examples, editable rule notes
and its own event ledger. It uses the same public fact representation and inquiry rule as the ladder,
with 20 scored calls per stream and a 60-second call deadline. Both receive four full-menu commissioning
queries and at most three optional queries per stream. Ordinary F1 labels follow only physically
completed actions. One physical attempt is allowed per job. All failures and timeouts remain in the
assigned-job denominator. Source, controller, model, normalisation, assignments and protocol were
frozen before evaluation. \Cref{app:continuing} gives the full protocol and accounting.

The primary pilot endpoint is accepted task completion: independently verified physical effect,
acceptable chosen action and service-verified original goal. Wrong dispatches, completed wrong actions,
execution failures, queries and setup costs are separate endpoints. A wrong dispatch may also fail
physically. The six purposively specified stream pairs provide descriptive evidence of integrated performance.
They do not support a powered superiority or non-inferiority claim, and jobs within a stream are
not independent replicates.

\paragraph{DROC adaptation follow-up.} We additionally evaluate a DROC plan-level
adaptation to verdict feedback \citep{zha2024droc}, retaining the ladder and custom frontier controls from the first pilot.
Its knowledge module distils its own feedback into an append-only constraint bank and selects
relevant entries; a separate planning call sees only those entries and the current scene.
The same service and six nominal streams are reused, with a new persistent robot/learner per stream.
The assignments and historical controls were already exposed, so this is a descriptive follow-up,
not a fresh confirmatory comparison. Both model stages share the first pilot's 60-second decision deadline;
interaction and physical budgets are unchanged. Binary verdicts replace language corrections,
and a fixed push replaces multi-step planning and skill-code generation. Low-level visual retrieval
is not exercised. This tests an explicit adaptation of the published plan-level mechanism, not a
full DROC reproduction; \cref{app:droc} details the deviations and accounting.

\paragraph{Fresh paired CarryOn comparison.} A second prospective pilot compares the
unchanged ladder with a CarryOn planning-loop adaptation on six fresh scene stream pairs,
again with 120 scored jobs per method. It imports CarryOn's native user/scene memory stores,
write/delete operations, context policies and scene-update validator. Sparse verdicts and
camera facts replace its native dialogue/image interface; shielding and recovery are outside
the common one-push contract. The scene assignments are new, while the six requirement cases
are inherited from the frontier comparison. Pair order alternates, and both methods run anew with independent
persistent services and identical interaction budgets. Model validity checks and the source freeze precede evaluation. A
60-second whole-decision deadline covers planning and any plan repairs; all failures are retained
under the prospective protocol. \Cref{app:carryon} specifies the retained native components,
interface changes and resource limits.

% End sections/05_setup.tex

% Begin sections/06_results.tex
\section{Results}
\label{sec:results}

\subsection{\RQ{1}: the ladder against correction memory}
% src: scope_ladder_asking_c7_20261107/C7_RESULT_v1.txt §1-3; C7_CORRECTION_v1.txt;
% results/violations_corrected_c7.json; counts derived from results/rows.jsonl (checks/recheck_c7_table.py).
The main confirmatory study tested the ladder on 150 fresh sites with four pre-registered conditions, all of which were met
(\cref{tab:c7}). Without asking, the ladder beat kNN memory on new contexts by $+0.074$ $[+0.042, +0.108]$
(bar $\geq +0.03$, CI $>0$), on repeated contexts by $+0.052$ $[+0.027, +0.079]$ (non-inferiority margin
$-0.03$; the CI is also above zero), and was not worse out of grammar ($+0.022$ $[+0.002, +0.044]$; margin
$-0.05$). With the same right to ask, it led on new contexts by $+0.051$ $[+0.023, +0.083]$, and on repeated
contexts by $+0.031$ $[+0.007, +0.056]$ (secondary), while asking fewer questions.

\begin{table}[t]
\centering\small
\caption{\RQ{1}, main confirmatory study: pattern A, mean per stream of 36 scored jobs over 150 fresh sites; each cell is new /
repeated contexts. Wrong actions are per 100 scored jobs, corrected after review (questions answered
``nothing acceptable'' had been counted as violations; \cref{app:ledger}). \napmem's threshold was set
transductively.}
\label{tab:c7}
\setlength{\tabcolsep}{4pt}
\begin{tabular}{lccccc}
\toprule
 & Net utility $\uparrow$ & Acceptable$/36$ & Wrong$/100$ $\downarrow$ & Held off$/36$ & Questions$/36$ \\
\midrule
\napmem             & 0.408 / 0.474 & 16.9 / 18.2 & 6.2 / 3.2  & 16.9 / 16.7 & -- \\
\knnmem             & 0.490 / 0.597 & 23.1 / 24.9 & 15.1 / 9.4 & 7.5 / 7.8   & -- \\
\ladder             & 0.564 / 0.650 & 24.7 / 25.9 & 12.4 / 6.9 & 6.8 / 7.6   & -- \\
\midrule
\knnmem\unc         & 0.532 / 0.651 & 25.2 / 27.1 & 13.0 / 7.1 & 6.1 / 6.4   & 2.79 / 2.15 \\
\ladder\unc         & \textbf{0.583} / \textbf{0.681} & 25.8 / 27.1 & 11.1 / 5.2 & 6.2 / 7.1   & 1.65 / 1.31 \\
\midrule
\ruleopen@true (ref.) & 0.653 / 0.690 & & & & \\
Oracle (ref.)  & 0.864 / 0.853 & & & & \\
\bottomrule
\end{tabular}
\end{table}

With matched asking, \ladder\unc{} took 14\% fewer wrong actions than \knnmem\unc{} on new contexts and 27\%
fewer on repeated ones. On repeated contexts both completed the same number of acceptable actions (27.1 of
36): there the gain comes from fewer wrong actions and fewer questions, not from more completions. A
pre-registered secondary analysis found a smaller benefit from asking for the ladder than for kNN memory:
the gain from asking under the ladder minus the gain under kNN memory is $-0.022$ on new and $-0.021$ on repeated
contexts, both with CIs just below zero. Asking therefore does not explain the ladder's lead.

\subsection{\RQ{2}: replication and component contributions}
% src: C6_RESULT_v1.txt §1-4; C8_RESULT_v1.txt §3e; decision_c6.json (Table 2 by checks/make_c6_ablation_table.py).
\begin{figure}[t]
  \centering
  \includegraphics[width=0.85\linewidth]{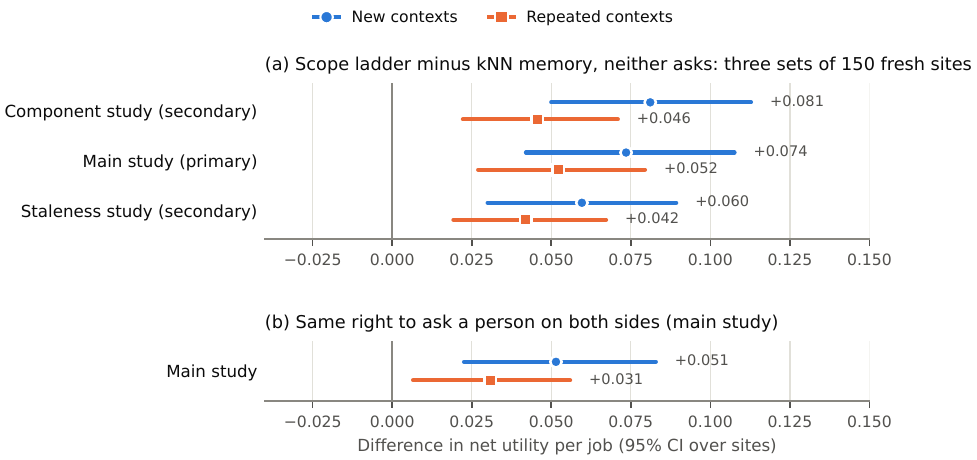}
  \caption{\RQ{2}. (a) The ladder's lead over kNN memory on three independent sets of 150 fresh sites,
  without asking: primary conditions in the main confirmatory study, and pre-registered secondary contrasts
  in the component and staleness studies.
  (b) The main confirmatory study with the same right to ask on both sides (new contexts: primary; repeated: secondary). Dots are
  means over sites, bars 95\% bootstrap CIs.}
  % src: decision_c6/c7/c8.json via figures/make_figures.py
  \label{fig:replication}
\end{figure}
The lead was measured on three independent sets of 150 fresh sites (\cref{fig:replication}): the component study (secondary)
$+0.081$ $[+0.050, +0.113]$ on new and $+0.046$ $[+0.022, +0.071]$ on repeated contexts; the main confirmatory study (primary) $+0.074$
and $+0.052$; the staleness study (secondary) $+0.060$ $[+0.030, +0.089]$ and $+0.042$ $[+0.019, +0.067]$.

\begin{table}[t]
\centering\small
\caption{\RQ{2}, component study: mean net utility per job over 150 fresh sites (71 with a rule
change for C-post, the jobs after the change). Columns are patterns: A fixed rule, O out of grammar, E 20\%
execution failures, C-post after an unannounced change. Bold: best learner per column. Generated from the frozen
decision file by script.}
\label{tab:ablation}
\begin{tabular}{lcccc@{\hspace{1.2em}}cccc}
\toprule
 & \multicolumn{4}{c}{N: every context new} & \multicolumn{4}{c}{R: about 62\% repeats} \\
\cmidrule(lr){2-5}\cmidrule(lr){6-9}
 & A & O & E & C-post & A & O & E & C-post \\
\midrule
% generated by checks/make_c6_ablation_table.py from decision_c6.json sha256 bdd49ec78fea2b02
\openlvl           & 0.509 & 0.207 & 0.402 & 0.173 & 0.562 & 0.241 & 0.450 & 0.234 \\
\ruleopen          & 0.551 & 0.245 & 0.437 & 0.282 & 0.621 & 0.276 & 0.496 & 0.332 \\
\mix               & 0.565 & 0.254 & 0.445 & 0.273 & 0.656 & 0.291 & 0.516 & 0.351 \\
\mixpj             & 0.558 & 0.254 & 0.438 & 0.287 & 0.660 & 0.303 & 0.521 & 0.361 \\
\ladder            & \textbf{0.573} & \textbf{0.263} & \textbf{0.451} & \textbf{0.291} & \textbf{0.699} & \textbf{0.324} & \textbf{0.554} & \textbf{0.372} \\
\knnmem            & 0.492 & 0.231 & 0.386 & 0.220 & 0.653 & 0.286 & 0.507 & 0.351 \\
\napmem            & 0.418 & 0.110 & 0.330 & 0.190 & 0.504 & 0.155 & 0.402 & 0.277 \\
\midrule
\ruleopen@true     & 0.682 & 0.299 & -- & -- & 0.732 & 0.306 & -- & -- \\
Oracle             & 0.861 & 0.478 & 0.688 & 0.880 & 0.862 & 0.466 & 0.692 & 0.890 \\
% sites: N {'A': 150, 'C': 71, 'E': 150, 'O': 150}, R {'A': 150, 'C': 71, 'E': 150, 'O': 150}
\bottomrule
\end{tabular}
\end{table}

\Cref{tab:ablation} decomposes the provider within the pre-registered component study, whose primary arm was \mixpj{}
and in which the ladder was a pre-registered secondary arm. The ladder is the best learner in every column.
Rule inference and the memory gate contribute in different regimes. On new contexts, rule inference
accounts for the gain:
\ruleopen{} alone beat kNN memory by $+0.059$ $[+0.024, +0.092]$, and the mixture beat it by $+0.073$
$[+0.041, +0.105]$ (both post hoc). With repeats, the same rule inference without memory \emph{lost} to kNN memory
($-0.033$ $[-0.063, -0.002]$, post hoc), as it had in the failed initial repeated-context study; adding memory to the mixture only matched kNN
memory ($+0.003$ $[-0.021, +0.028]$, post hoc), and the gate supplied the lead (0.699 against 0.656 for the
mixture, descriptive; $+0.038$ $[+0.028, +0.050]$ over \mixpj, post hoc). Per-job evidence (\mixpj) changed utility little relative to the global mixture ($-0.007$ new, $+0.004$ repeated). As a pre-registered primary,
\mixpj{} met all three of the component study's conditions: $+0.066$ $[+0.034, +0.098]$ over kNN memory on new contexts, $+0.007$
$[-0.018, +0.034]$ with repeats (non-inferior at margin $-0.03$), and $+0.023$ $[+0.003, +0.043]$ out of grammar.
All learners stay well below the oracle ($0.861$ new, $0.862$ repeated in pattern A): replacing noisy probe facts
with true simulator facts (\ruleopen@true) raises \ruleopen{} by $+0.131$ on new contexts ($0.551$ to $0.682$)
and $+0.112$ on repeated ones ($0.621$ to $0.732$), while a further $0.179$ (new) and $0.130$ (repeated) gap to
the oracle reflects partial bandit feedback and held-out rule families.

\subsection{\RQ{3}: other rule learners}
% src: C10_RESULT_v1.txt; D11_RESULT.txt; D16_RESULT.txt; D18_RESULT.txt (corrected 8 Nov); D20_RESULT.txt.
\begin{figure}[t]
  \centering
  \includegraphics[width=0.85\linewidth]{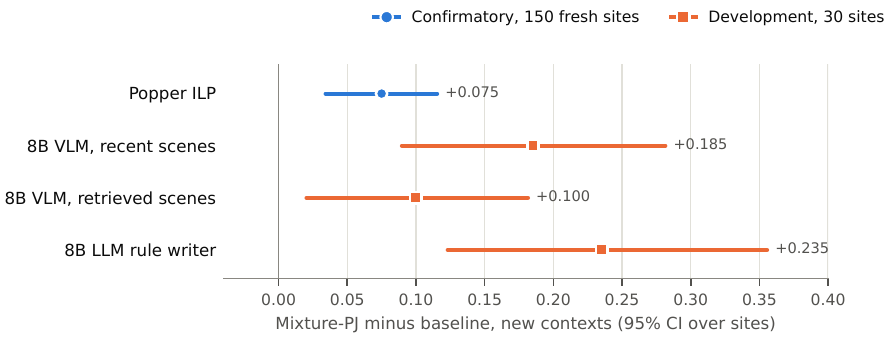}
  \caption{\RQ{3}. \mixpj{} minus other rule proposers and learned models on new contexts, pattern A. Popper ILP
  (one configuration; tie-breaking seed not controlled) is a confirmatory result on 150 fresh sites; the
  8B VLM and LLM comparisons are development results on 30 sites.}
  % src: decision_c10.json, summary_d16.json, summary_d11.json
  \label{fig:baselines}
\end{figure}

\paragraph{Popper (confirmatory).} With the same perceptual vocabulary, \mixpj{} beat
Popper on new contexts by $+0.075$ $[+0.034, +0.115]$, and \ruleopen{} beat it by $+0.064$ $[+0.023, +0.105]$.
On repeated contexts, Popper alone also lost to \knnmem{} by $-0.047$ $[-0.088, -0.007]$ ($0.545$ vs $0.592$),
consistent with the repetition deficit of un-gated grammar inference. When Popper replaced the grammar as the
ladder's rule level, the ladder improved on Popper alone by $+0.072$ $[+0.049, +0.096]$ new and $+0.087$
$[+0.059, +0.116]$ repeated: the ladder's arbitration is not tied to one rule proposer. Both primary effects were
about two thirds of their development size, and Popper averaged $1.00\,\mathrm{s}$ per learning call across
$29{,}045$ calls. This comparison uses one Popper configuration over a restricted predicate vocabulary
(\cref{app:settings}). A later development study found that Popper's tie-breaking between equally good
programs depends on the Python hash seed, which this confirmatory study did not control. Its intervals are therefore
conditional on the realised tie-breaks; a fixed-seed replication has not been run.

\paragraph{Learned models (development only).} On 30 development sites, an 8B VLM given earlier scenes and their
verdicts trailed \mixpj{} on new contexts by $0.185$ (recent scenes) and $0.100$ (retrieved similar scenes).
The 8B LLM rule writer trailed by $0.235$ on new contexts and by $0.089$ out of grammar; its rules were
mostly too narrow. A frontier model writing rules performed better on six streams; out of grammar, it was ahead
of our learner on two streams and tied on one. The continuing-service pilot evaluates a
frontier agent combining retrieved instances, editable rules and refinement (\cref{sec:integrated});
it uses a separate physical task population and does not extend these decision-level contrasts.

\subsection{\RQ{4}: rule changes}
% src: C7_RESULT_v1.txt §3b, §4; C8_RESULT_v1.txt.
After an unannounced rule change with repeated contexts, the ladder trusts memories recorded under the old
rule: without asking, its post-change utility in the main confirmatory study was below kNN memory's (0.298 vs 0.317); with asking the
difference was $+0.001$ $[-0.038, +0.039]$, inconclusive. On new contexts, where nothing stale is reused,
\ladder\unc{} led \knnmem\unc{} after the change by $+0.167$ $[+0.098, +0.234]$ (70 sites). The staleness study tested the
staleness rule on the ladder's memory. It improved post-change utility with repeats by $+0.027$
$[+0.015, +0.040]$ (68 sites with a rule change) and changed nothing when the rule was fixed (exactly $0.000$ in
patterns A and O: the rule never fired). The pre-registered criterion required a positive CI without a minimum practical effect.
The improvement closes about 5\% of the post-change gap to the oracle (0.351 to 0.378, oracle 0.883).

\subsection{\RQ{5}: integrated continuing-service comparison}
\label{sec:integrated}
\begin{figure}[t]
\centering
\includegraphics[width=\linewidth]{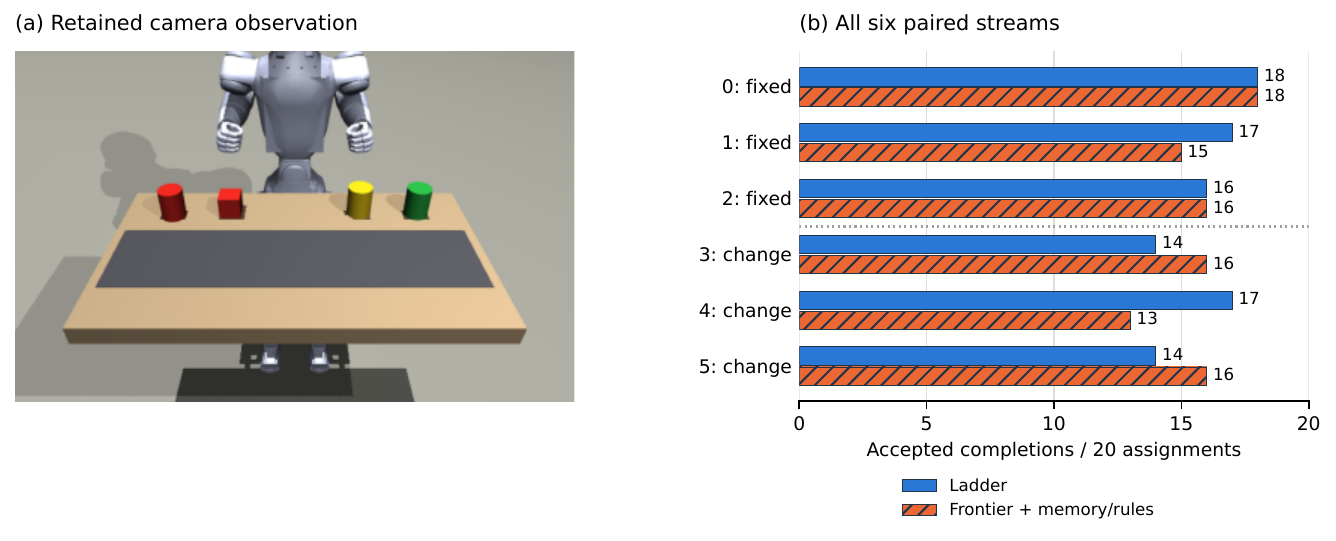}
\caption{Continuing-service pilot. Left: a retained camera observation from the fixed-pelvis R1
tabletop scene. Right: accepted completions over all 20 scored assignments in each paired stream.
Streams 0--2 have fixed requirements; streams 3--5 change requirement. Object layouts vary, and
external setup is logged between jobs. This is a six-pair simulation pilot using one push skill.}
\label{fig:integrated}
\end{figure}

% src: continuous_ladder_20261110/RESULTS_v1.md and results_v1/{summary,extended_audit}.json.
The pilot completed all 288 assignments within its frozen time budget, with no omissions or selective
reruns. Both methods used the same continuing service, with one persistent native session per
method/stream. \Cref{tab:integrated} reports every scored assignment; \cref{fig:integrated} shows a
camera observation and the paired outcomes.

\begin{table}[t]
\centering\small
\caption{Continuing-service pilot: all 120 scored assignments per method. Optional queries exclude
24 commissioning queries per method. Wrong dispatches and execution failures overlap once per
method; injected failures are included in execution failures.}
\label{tab:integrated}
\begin{tabular}{lrr}
\toprule
Outcome & Ladder & Frontier + memory/rules \\
\midrule
Accepted task completion & 96/120 (80.0\%) & 94/120 (78.3\%) \\
Verified physical completion & 108 & 102 \\
Wrong actions dispatched & 13 & 9 \\
Completed wrong actions & 12 & 8 \\
Execution failures & 12 & 11 \\
\quad Including injected interruptions & 2 & 2 \\
Model decision timeouts & 0 & 7 \\
Optional full-menu questions & 6 & 11 \\
Abstained / unstarted / unresolved & 0 / 0 / 0 & 0 / 0 / 0 \\
\bottomrule
\end{tabular}
\end{table}

\paragraph{Completion, errors and adaptation.} The ladder completed two more accepted jobs and asked
five fewer optional questions; the frontier made four fewer wrong dispatches but incurred seven
decision timeouts. Across 30 post-change jobs per method, the ladder accepted 21 and the frontier 23,
with nine versus four wrong dispatches. Both accepted 22/24 revisited jobs. The descriptive comparison favours the ladder on completion and query count, and the frontier on
wrong dispatches and post-change completion. The frontier's latency budget affects completion.
Moreover, some old and new rules permit the same action, so post-change success need not imply
full rule recovery.

\paragraph{Execution and acceptability.} Physical completion alone would count 108 ladder jobs
as successful, although only 96 satisfied the hidden requirement and original goal. All 23 scored
execution failures remained failures; four were scheduled interruptions, which confirmed stops and
supplied no F1 labels. The other 19 stopped short of the required clear zone. These records verify that the service reported execution and acceptability separately and
withheld labels after failed execution. Both providers use the
harness, so the experiment does not isolate a causal benefit over execution without those contracts.

\paragraph{Intervention and compute costs.} Each method also received 24 commissioning queries
and required 144 external scene setups. Total labels, including commissioning, were 208 versus 217.
These are simulated oracle events rather than measured human effort. Under the declared utility
proxy (accepted minus completed-wrong minus $c$ times optional questions), the ranking crosses at
$c=0.4$ (\cref{tab:cost}). Scored-job wall time was 11.02 versus 71.93 minutes; the latter includes
the configured CLI/model latency. These timings are not physical-robot throughput estimates.

\paragraph{Connection to the earlier failure case.} The initial closed-loop demonstration used one 12-job Panda stream and did not
recover from a white-to-blue placement-rule change; its fresh-per-job simulation and perception
failures differed from this pilot (\cref{app:c9}). The pilot ran through the continuing service and completed some post-change jobs, but reliable
change recovery remains unresolved. Post-hoc gate diagnostics do not support attributing all
post-change errors to stale memory (\cref{app:continuing}).

\paragraph{DROC adaptation follow-up.} The DROC plan-level verdict adaptation
accepted 93/120 scored jobs, compared with the frontier pilot's retained 96/120 for the ladder and 94/120 for the
custom frontier (\cref{tab:droc}). It made 13 wrong dispatches, used 12 optional questions and had
two abstentions, 14 execution failures and no model errors. At question cost $c=0.5$, its utility
was 76, versus 81 and 80.5 for the historical controls. It accepted 19/30 post-change jobs and
22/24 revisits. All 144 assignments were retained and independently audited (\cref{app:droc}).
The comparison tests an adaptation of a published correction-learning mechanism. Its changed
interface, exposed streams and non-concurrent controls preclude a native-DROC or general
superiority claim. Neither completion difference reached the protocol's descriptive
five-percentage-point reference for motivating a larger study.

\paragraph{Fresh paired CarryOn planning comparison.} On six new scene stream pairs, the
unchanged ladder and CarryOn planning adaptation accepted 96/120 and 90/120 scored jobs,
respectively (\cref{tab:carryon}). Unlike the DROC follow-up, both arms were run on the fresh assignments with
alternating order and independent persistent services. CarryOn's native user/scene memory stores,
CRUD operations and context policies are executed; its planning loop is adapted to the common
camera-fact, one-push and sparse-verdict interface (\cref{app:carryon}). The requirement cases
themselves are inherited from the frontier comparison, and these results are not pooled with either earlier pilot.

The ladder asked seven optional questions versus seventeen, but made twelve wrong dispatches
versus nine. Dispatch coverage was 118/120 versus 109/120: both methods abstained twice, and
CarryOn incurred nine decision timeouts. Both had eleven scored execution failures, including
two injected interruptions each. At $c=0.5$, utility was 81.5 versus 73.5. The six-job completion
gap is exactly five percentage points, meeting the protocol's descriptive practical reference;
six stream pairs do not establish statistical superiority. Nor can the gap be attributed solely
to correction learning, because completion includes the configured model deadline.

The relative completion rates differ between fixed-requirement and post-change jobs. The ladder accepted 56/60 jobs on fixed-requirement
streams versus CarryOn's 49/60, but after changes it accepted 17/30 versus 22/30, with nine versus
five wrong dispatches. CarryOn also accepted 20/24 revisits versus 18/24. All 288 assignments,
including model errors and failed executions, are retained and independently audited. This fresh paired comparison evaluates native memory mechanisms through the shared contract.
CarryOn's full visual planning, shielding and recovery system remain outside its scope.

% Begin sections/tab_carryon.tex
\begin{table}[ht]
\centering\small
\caption{Fresh paired CarryOn planning-loop comparison: all 120 scored assignments per method. The methods share the continuing service; CarryOn uses native memory components with the documented verdict-interface adaptation. Wrong dispatches can overlap execution failures.}
\label{tab:carryon}
\begin{tabular}{lrr}
\toprule
Metric & Ladder & CarryOn-plan \\
\midrule
Accepted task completions & 96 & 90 \\
Actions dispatched & 118 & 109 \\
Physical completions & 107 & 98 \\
Wrong actions dispatched & 12 & 9 \\
Completed wrong actions & 11 & 8 \\
Execution failures & 11 & 11 \\
Injected failures (included above) & 2 & 2 \\
Provider errors & 0 & 9 \\
Optional full-menu questions & 7 & 17 \\
Abstentions & 2 & 2 \\
Unstarted assignments & 0 & 0 \\
Unresolved jobs & 0 & 0 \\
Utility ($c=0.5$) & 81.5 & 73.5 \\
\bottomrule
\end{tabular}
\end{table}

% End sections/tab_carryon.tex

% End sections/06_results.tex

\FloatBarrier
% Begin sections/07_discussion.tex
\section{Discussion}
\label{sec:discussion}

\subsection{What did not work}
\label{sec:negative}
% src: STUDY_INDEX.md §2; individual RESULT files named in Appendix D.
The negative results help explain the ladder's design (\cref{app:negative}).
\emph{Structured rule prior:} learned scope embeddings failed to beat simple
baselines in two pre-registered studies, motivating an explicit hypothesis space over
perceived facts that subsequently met its offline evaluation criteria. \emph{Specialist memory gate:} rule inference without memory lost to kNN
memory on streams with repeats ($-0.031$ on the pooled primary endpoint, failing the pre-set criteria) and missed its criterion without
repeats, while a global mixture still lost to kNN on repeats in development; among these variants, gating near-identical
instances to memory preserved repetition performance while keeping rule generalisation on new contexts.
\emph{Perception post-processing:} downstream patches on the learner's side (probe
uncertainty, depth localisation, an instance detector) recovered at most $+0.011$ of the $+0.131$
true-fact gap on new contexts. This suggests that improving the frozen visual representation may be
more useful than the tested changes to how the rule learner consumes it. \emph{Uncertainty-based asking:} a scope-aware question
selector did not beat the plain uncertainty band in development, and heuristic post-error asking triggers did not help
after rule changes. \emph{What remains unsolved:} rule-level recovery after unannounced changes (a
recent-window switch missed its primary criterion in development), adding an 8B LLM as an extra rule level (which
reduced reliability in development), and applying the ladder to skill selection where NAPPING already suffices. Three studies
failed their own validity checks and are not results. Before any of this, a line of repair-memory studies in
physics simulation could not separate the memories it compared (\cref{app:repair}): with two candidate actions
and a handful of distinct situations, there was little structure in \emph{where} a lesson applied. This motivated the study of reach when choosing actions under unwritten site requirements.

\subsection{Future work}
\label{sec:limits}
\textbf{Physical robots and human feedback.} Extending the simulation studies to physical robots
would test CRAFT across more skills and with noisy or delayed feedback. The current trials use
correct simulated feedback, one fixed-pelvis push skill and external setup. Human studies should
compare full-menu labels with single yes/no verdicts and measure time and effort directly, beyond
query counts.

\textbf{Independent requirements.} Larger prospective studies should use requirements authored
independently of the learner's grammar. The current grammar overlaps with the generator's rule
families, and the integrated comparisons reuse six hand-specified cases. Independently authored
requirements would provide a stronger test of transfer to unfamiliar sites.

\textbf{Learning scope and revising rules.} Learning or calibrating the memory gate could extend
its current fixed-threshold design. Integrated ablations would test whether the decision-level
mechanism findings carry through to execution. Revision after unannounced changes deserves
particular attention: stale memories and confident rule predictions can persist, and the frontier
agent and CarryOn adaptation performed better after changes than their paired ladder controls.

\textbf{Broader comparisons.} Prospective paired comparisons should extend the small descriptive
studies across frontier configurations and resource budgets. Evaluating fuller DROC and CarryOn
systems would complement the present interface adaptations (\cref{app:droc,app:carryon}); fresh
paired controls would also replace DROC's exposed streams and historical controls. A separate
service ablation would help distinguish the harness's contribution from that of the decision
provider, since both sides currently use the shared service.

\section{Conclusion}
\label{sec:conclusion}
Learning unwritten requirements from sparse verdicts requires deciding how far a correction should
reach. CRAFT makes that decision a replaceable provider under a common evidence and feedback contract.
The scope ladder gives memory precedence on near-identical labelled situations and combines rule,
open and memory predictions elsewhere using their past predictive evidence. Controlled studies show gains over correction memory on new and repeated
synthetic situations; component comparisons explain the value of arbitration within those populations.
A continuing-service pilot connects the learner to camera-grounded execution and a frontier
memory/rule agent. Accepted completion is similar, but the ladder asks fewer questions and the frontier makes fewer
wrong dispatches; all failures remain in the comparison. A DROC plan-level verdict adaptation
adds descriptive evidence on the exposed streams, with 93/120 accepted jobs and higher query cost;
it does not establish performance against native DROC. A fresh paired CarryOn planning adaptation
accepts 90/120 versus the ladder's 96/120 and asks more questions, while making fewer wrong
dispatches and performing better after changes. Its nine timeouts affect completion, and the changed interface limits generalisation to native CarryOn. The next scientific test is a larger
prospective comparison on independently specified requirements, followed by broader skills and
physical-robot feedback. Reliable revision after unannounced changes remains an open problem.

\paragraph{Code availability.} Code will be released at
\url{https://github.com/zp312/CRAFT}; the repository currently provides a project overview.

% End sections/07_discussion.tex

\bibliographystyle{plainnat}
\bibliography{refs}

\clearpage
\appendix
% Begin sections/A_harness.tex
\section{Harness details}
\label{app:harness}
% src: ARCHITECTURE_v1.txt (4 Nov 2026); HARNESS_TASK_RUNNER_E1_REPORT_v1.txt; K1 v2 / K2 v1 contracts
% (COORDINATION_v1.txt, hashes 8da61a0b..., 6787ddad...).

This appendix expands \cref{sec:problem}; \cref{fig:arch} shows the path of one decision.

\subsection{Authority}
Authority is deliberately narrow, and each role belongs to exactly one component:
\begin{itemize}
  \item the \emph{deployment service} admits and sequences jobs;
  \item the \emph{task runner} owns one job's goals, verified progress and budgets;
  \item \emph{planner and adaptation providers} propose; they cannot dispatch physical commands;
  \item the \emph{supervisor} owns dispatch, physical custody, deadlines and cancellation;
  \item the \emph{native backend} owns motor commands and its local stop behaviour;
  \item \emph{independent verifiers} supply the evidence for progress, which the task runner records.
\end{itemize}
Declared site rules live in a rule store and apply equally to every provider. A learned proposal cannot
relax an authoritative constraint, and a physical success cannot remove a prohibition.

\subsection{Three loops}
\emph{Native control} runs independently of task reasoning and network services and keeps each
controller's own rate and stop behaviour; a cancellation acknowledgement is not treated as proof that
the robot has stopped. \emph{One bounded job} reads fresh evidence, asks the fixed planner for eligible
actions, obtains one adaptation proposal, dispatches at most one physical operation, verifies its
actual effects, and records progress and factual experience; success is recorded only after the
original goal and required constraints are checked. \emph{Continuing deployment} admits the next job
without resetting physical state or erasing experience.

\subsection{Contracts that make providers comparable}
The records below are the architecture's contracts (\texttt{ARCHITECTURE\_v1}); the implemented code uses
narrower records, mapped in \cref{sec:implstatus}. A provider sees a \texttt{DecisionContext}: fresh public evidence, observed site cues, the original and
remaining goals, declared constraints, the qualified actions and the provider's own factual history. It
excludes private condition identifiers, future outcomes and hidden rule revisions. It returns a
\texttt{DecisionProposal} (execute, observe, wait, ask for help, finish) with evidence references, a
reason and the cost it charges. After execution it receives a \texttt{FeedbackRecord} in which the
\emph{execution effect}, the \emph{acceptability} of the action under the site's requirements and the
\emph{whole-task} verdict are separate fields, each TRUE, FALSE or UNKNOWN. A generic failure bit would
not tell the learner whether the motor, the destination or the remembered rule was wrong; separating them lets a requirement learner ignore execution failures and avoid learning false
rules from them.

The feedback contract fixes what a provider may learn. Under the level used throughout this paper
(F1), the provider learns the acceptability of the action it actually executed and nothing about
actions it did not take; abstentions and failed executions return no acceptability verdict. Asking a
person is a separate, charged channel (\cref{sec:problem}).

\subsection{Information-boundary tests}
For the selector modules in the initial repeated-context, initial new-context and component studies
(the grammar, open, kNN and mixture providers and the gate), three tests ran before any test output: a \emph{canary} test (private fields replaced by sentinel values must leave
every decision bit-identical), a \emph{static import} check (provider modules cannot import the label
generator, rule evaluator or bank loader), and a \emph{prefix-invariance} test (decisions up to job $t$
must not change when everything after $t$ is randomised). Later providers had study-specific tests only:
The confirmatory Popper comparison checked class bindings and equivalence on smoke runs. The
development study of the corrected Popper configuration checked site separation and the ILP configuration. The ILP provider's module imports code that also loads evaluator exports, so the static
import check does not cover it. We found no evidence of label leakage, but the coverage of these tests is
not uniform across providers.

\subsection{Implementation and evaluation coverage}
\label{sec:implstatus}
% src: ARCHITECTURE_v1.txt §4; src/robot_agent_harness/task_contracts.py; experiments/craft_comparison_driver_v1_20261106/
% public_contracts.py; scope_closed_loop_c9_20261107/run_c9.py; HARNESS_TASK_RUNNER_E1_REPORT_v1.txt ("542 passed").
\begin{itemize}
  \item \emph{Implemented as bounded prototypes:} the supervisor, the skill registry and native simulation
    backends; a reusable task runner and an in-memory deployment service (records \texttt{TaskSnapshot},
    \texttt{TaskDecision}, \texttt{ExecutionOutcome}), which pass 542 software checks. These checks assess the software lifecycle; they do not estimate robot reliability.
  \item \emph{Partial:} the rule store and the shared experience service. A CPU comparison driver
    implements the public decision boundary (\texttt{PublicDecisionContext}, \texttt{Decision},
    \texttt{PublicJobFeedback}); its feedback record carries the F1 verdict but not the three separate
    TRUE/FALSE/UNKNOWN fields described above.
  \item \emph{Not evaluated here:} durable restart recovery, a human operator interface and
    autonomous physical-world deployment.
  \item \emph{What the studies used:} the confirmatory and development studies ran providers in a study
    driver at the decision level (\cref{sec:protocol}). The initial closed-loop demonstration launched the frozen
    skill layer's runner in a fresh simulator process for each job and used the comparison driver's F1
    projection; it did not exercise the task runner or deployment service over a continuing deployment.
  \item \emph{Continuing-service frontier comparison:} both providers used the unchanged task runner, supervisor and deployment
    service, retaining one native session through each 24-job stream. Its six contract checks and
    post-run audits cover original-learner parity, public projection, past-only histories, exactly-once
    learning, failure-label suppression and persistent state (\cref{app:continuing}).
\end{itemize}

% End sections/A_harness.tex

% Begin sections/B_skills.tex
\section{A frozen, perception-grounded skill layer}
\label{app:skills}
% src: perception_grounded_skills_v1_20261105/MILESTONE_RESULT_v1.txt (5 Nov 2026);
% qualified manifest SHA-256 87ef9826...a8c11c.

Adaptation methods are compared above one released skill layer that none of them can change. The
layer serves a MuJoCo tabletop scene from the LIBERO suite (\texttt{3-4\_104\_1}) with a simulated Franka Panda
arm, a $512\times512$ \texttt{agentview} RGB-D camera and $20\,\mathrm{Hz}$ operational-space control: a known
black bowl and white and blue
plates. It supplies five skills (\texttt{move\_to\_target}, \texttt{close\_lift}, \texttt{reorient\_held},
\texttt{transport}, \texttt{lower\_release\_retreat}). Each uses the current RGB-D image, the camera
calibration and robot proprioception, and issues bounded seven-dimensional operational-space intents
(translation and rotation increments capped at 10\,mm and 50\,mrad), under shared per-episode ceilings
of 560 actions and 1{,}200 perception queries. Perception is a disclosed known-appearance colour and
shape model with current depth geometry; when an object is not found, the skill abstains.

\paragraph{Qualification.} The layer was frozen and then qualified on ten prospective episodes: five
matched initial states, each requested to the white and to the blue plate (3{,}074 simulator actions and
4{,}132 grounding queries in total). Eight completed the whole
controller sequence and passed independent native checks on placement, no gripper contact, object speed
and arm stop. The two failures shared one initial state, in which the bowl-grounding query abstained
before any controller action; both stay in the denominator. Eight of ten is a small-sample screen, not
a reliability estimate. The release seals 5{,}947 input files; evaluator poses and contact truth were
recorded separately and never reached a controller or provider.

\paragraph{What the layer does not cover.} Other objects, open-vocabulary recognition, broader
workspaces and real robots are not qualified. In the initial closed-loop demonstration (\cref{app:c9}), the same
bowl-grounding abstention is the main execution failure.

\paragraph{Separate R1 push adapter.} The continuing-service pilot uses a different frozen skill:
fixed-pelvis R1 tabletop pushing with camera-derived object positions, inverse kinematics and PD
control. Its separate prospective screen passed 14/16. This does not expand the Panda qualification
above; its domain, completion checks and retained failures are specified in \cref{app:continuing}.

% End sections/B_skills.tex

% Begin sections/C_repair.tex
\section{Lessons from repair-memory studies}
\label{app:repair}
% src: output/adaptation_research/HARNESS_PANDA_PROSPECTIVE_MEMORY_REPORT_v2.txt;
% HARNESS_Q2_COMPARISON_REPORT_v1.txt (protocol SHA-256 6114ede8...); HARNESS_BENEFIT_MEMORY_REPORT_v1.txt;
% HARNESS_REPAIR_APPLICABILITY_REPORT_v1.txt. Revised after review (review/REVIEW_v1.md M8, M9).

CRAFT's first research question was which repair to reuse after a skill failure. In a physics-simulated
container-emptying task with a Panda arm, after a slippery object fails under the nominal grasp, the
harness can continue or perform a rim re-grasp. Within each study, every method received the same
observations (six fixed grasp features), candidate actions and prescribed discovery attempts, and could
learn only from its own executed outcomes. Episodes were reset between encounters.

\paragraph{No memory was established as better.} In a prospective comparison frozen before its run (eight
streams with six prescribed discovery trials each, 48 autonomous situations, 672 physics executions),
NAPPING completed 40 of 48 tasks and three-nearest-neighbour memory 39 of 48, against an evaluator
ceiling of 43; the exploratory paired interval includes zero, which shows no advantage but not
equivalence. In a second frozen comparison on a withheld condition (four sequences, two prescribed
discovery attempts each), post-change experience mattered (NAPPING 16 of 16 post-discovery slippery
queries, versus 0 of 16 for the same memory frozen before the change), and NAPPING and kNN tied exactly
on the post-discovery subset (32 of 32 each, no harmful interventions).

\paragraph{Learned applicability did not improve completion.} A learned utility predictor and an applicability gate
with support, counterexample and margin checks were tried twice. In a small prospective engineering run
(12 matched autonomous cases; 168 policy and 24 evaluator executions) the applicability gate completed 8
of 12 tasks, against 11 for NAPPING and for the utility memory. In a later grouped development replay
(retrospective, no new physics, 48 situations) the applicability gate completed 32 of 48 and a
benefit-gated memory 34 of 48, against 41 for NAPPING: their evidence gates withheld useful repairs. The
ungated benefit memory also completed 41 and did not improve on NAPPING. We did not run a larger
prospective comparison.

\paragraph{Implications for task design.} With two candidate actions and a handful of distinct situations,
there was little structure in \emph{where} a lesson applied, and the comparisons did not distinguish
the memories. Equal information and retained failures made the comparisons interpretable, but did
not make this task informative about correction reach. We therefore studied the choice among
admissible actions under unwritten site requirements.

% End sections/C_repair.tex

% Begin sections/D_ledger.tex
\section{Study overview}
\label{app:ledger}
% Full chronology and protocol hashes: records/FULL_STUDY_LEDGER_20261112.tex.
The evidence comprises confirmatory decision-level studies, development studies and
descriptive simulation pilots. \Cref{tab:ledger} summarises every confirmatory study's
outcome. Their protocols were frozen before outcome data existed; each online study
used 150 fresh sites and was evaluated once. The scope-embedding and grammar-prior
studies used existing scene banks offline.

\begin{table}[htbp]
\centering\small
\caption{Confirmatory studies and their pre-specified outcomes.}
\label{tab:ledger}
\begin{tabular}{@{}>{\raggedright\arraybackslash}p{0.35\linewidth}>{\raggedright\arraybackslash}p{0.59\linewidth}@{}}
\toprule
Study & Outcome \\
\midrule
Learned scope embedding, two protocols & Neither met its criteria against simple baselines. \\
Offline grammar prior & Met its criteria on perceived facts. \\
Initial repeated- and new-context selectors & Both missed their primary criteria; the mixture passed its own condition on new contexts. \\
Component and main studies & Both met their criteria; component and ladder results are reported in \cref{sec:results}. \\
Staleness study & Met its criteria for revision of correction memory. \\
Popper comparison & Met both criteria for comparing the ladder with ILP and for using ILP inside the ladder. \\
\bottomrule
\end{tabular}
\end{table}

Development studies used previously examined sites, with written plans but no
confirmatory claim. They informed configuration and mechanism choices; unsuccessful
variants and invalid studies are summarised in \cref{app:negative}.
The continuing-service frontier and fresh paired CarryOn comparisons are descriptive
pilots. The DROC follow-up uses exposed assignments and historical controls.
All assigned jobs, including execution failures and model timeouts, remain in their
respective records (\cref{app:continuing,app:droc,app:carryon}). The earlier closed-loop
demonstration and repair-memory studies are reported separately (\cref{app:c9,app:repair}).

\paragraph{Corrected secondary counts.} Questions answered ``nothing acceptable''
were initially counted as violations in the main confirmatory study. The counts in
\cref{tab:c7} were recomputed from frozen records; utilities, decisions and the primary
outcome are unchanged. The full chronological ledger, protocol hashes and correction
history are retained in the project archive.

% End sections/D_ledger.tex

% Begin sections/E_settings.tex
\section{Settings}
\label{app:settings}
% src: RESULTS_NARRATIVE_v1.md §2; C5_DEVIATIONS K4, K7; C7_PROTOCOL §2; C8_PROTOCOL §2-3.
\begin{figure}[htbp]
  \centering
  \includegraphics[width=\linewidth]{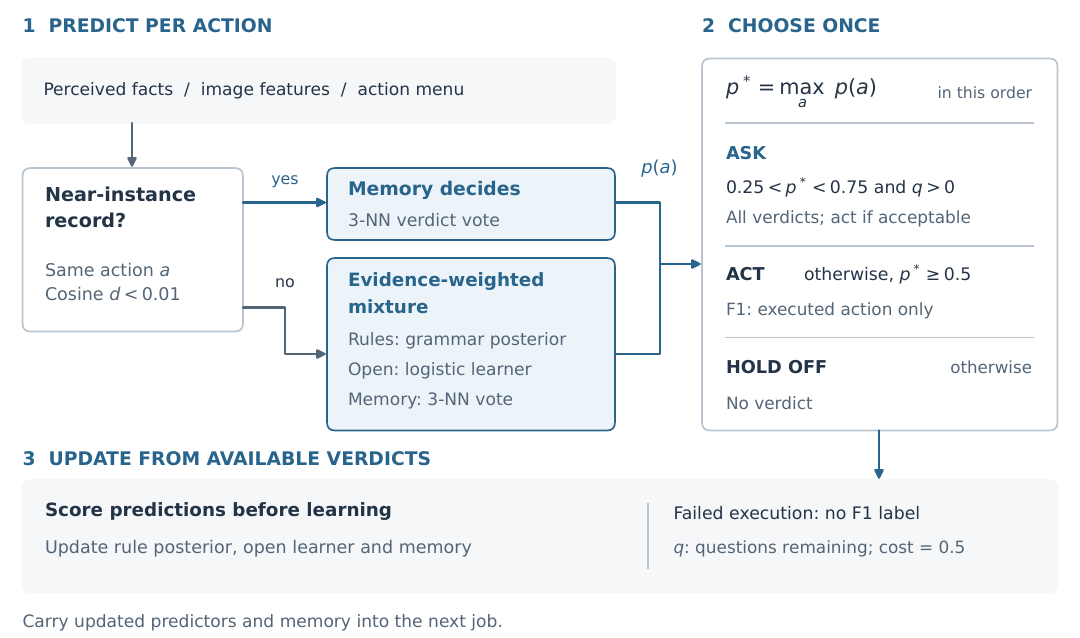}
  \caption{One scored job under the scope ladder with asking (\ladder\unc, as in the main confirmatory study). For each action, memory
  decides alone if the robot has been told about that action in a near-identical situation; otherwise the
  evidence-weighted mixture decides. A question reveals every action's verdict; without a question, F1
  reveals only the completed action's verdict. Failed executions and abstentions supply no F1 verdict.
  The decision and asking rules are the same for every provider.}
  \label{fig:ladder}
\end{figure}

\paragraph{Arm names.} We use the method names defined in \cref{sec:protocol} throughout the paper.
The suffix \unc{} marks a method with the right to ask.

\begin{itemize}
  \item \textbf{Backbone and probe.} Frozen SmolVLA vision tower (16$\times$16 pooled vision tokens,
    8$\times$8 connector tokens, 16$\times$16 pooled depth); two cross-attention layers with one query per
    catalogue object; trained on separate scenes only; one checkpoint (seed 0) online.
  \item \textbf{Grammar.} Seven-colour vocabulary; spatial ties within 1\,mm kept as a set; occupancy
    radii $\{0.06, 0.08, 0.10\}$\,m; composite thresholds 0.25--0.95\,m in 0.05\,m steps; median 1{,}807
    rules per site (range 1{,}404--2{,}731). Prior uniform over families, then within family.
  \item \textbf{Posterior.} Label noise $\varepsilon=0.05$; fixed-share rate $\eta=0.002$ (selected on
    development sites).
  \item \textbf{Open level.} L2 logistic regression ($C=1$), hard 0/1 output, $0.5$ with no records;
    prequential score with $\varepsilon_{\text{open}}=0.25$.
  \item \textbf{Memory.} Standardised, L2-normalised pooled features; cosine distance; $k=3$,
    weights $1/(d+10^{-6})$, no recency decay inside the mixture. \knnmem{}: $k=3$, $\gamma=1.0$ (N);
    $k=5$, $\gamma=0.95$ (R), chosen on development sites.
  \item \textbf{Mixture.} Prior $(0.6, 0.2, 0.2)$ for (grammar, open, memory); \ruleopen{} uses $(0.8, 0.2)$
    without memory. Memory score clipped to $[0.05, 0.95]$ for evidence only.
  \item \textbf{Gate.} ``Seen'' if a record of the same action lies within cosine distance $0.01$.
  \item \textbf{Decision.} Argmax, first action on ties; hold off if $\max_a p(a)<0.5$.
  \item \textbf{Evidence (prequential, per label, before the predictor learns it).} Grammar:
    $\log\sum_h w_h\,\mathrm{lik}_h$ with $\mathrm{lik}_h=1-\varepsilon$ if rule $h$ labels the action as observed,
    else $\varepsilon$. Open level: $\log 0.5$ with no records, else $\log 0.75$ if its hard prediction is right and
    $\log 0.25$ if wrong. Memory: $\log q$ or $\log(1-q)$ with its vote $q$ clipped to $[0.05, 0.95]$ (the unclipped
    vote is used for prediction). Order within a job: memory is scored before it stores the job's labels,
    then grammar and open level are scored and updated; labels of a fully labelled job are processed one
    action at a time; fixed-share forgetting is applied before each job after the first.
  \item \textbf{\napmem{} (NAPPING-derived correction memory).} Distance threshold $=$ factor $\times$ the median
    within-site distance between perceived-fact vectors, computed on the test sites' facts before the
    streams ran (no labels): factor 1.5 with repeats, 4.0 without; abstain when no record matches.
  \item \textbf{ILP (Popper, selected configuration).} Head \texttt{acceptable(job, action)}; a restricted set of
    body predicates computed from the perceived facts, with a single occupancy radius of 0.10\,m; at most 2
    clauses, 4 variables and 4 body literals; exact search first, noisy search as fallback; 10\,s timeout per
    call. Chosen among three configurations on tuning sites in the corrected Popper development study.
  \item \textbf{Asking.} Budget 3 per stream; cost $0.5$; ask iff $0.25<\max_a p(a)<0.75$; the answer
    gives every action's verdict in the context.
\end{itemize}

% End sections/E_settings.tex

% Begin sections/F_negative.tex
\section{Additional negative results}
\label{app:negative}
% Full earlier text: records/FULL_NEGATIVE_RESULTS_20261112.tex.
These results supplement the design discussion in \cref{sec:negative}. Unless
identified as confirmatory in \cref{app:ledger}, they are development findings.

\paragraph{Memory and perception.} Rule inference without memory had a utility
difference of $-0.031$ against kNN on the pooled primary endpoint of patterns A and C
with repeats. Without
repeats, \ruleopen{} missed its criterion, although \mix{} passed its own condition
($+0.051$) within the failed study. Perception post-processing using probe uncertainty,
depth localisation and an instance detector recovered only $+0.011$, $-0.002$ and
$+0.002$, respectively, of a gap of about $0.13$.

\paragraph{Asking and rule changes.} Scope-aware asking did not beat uncertainty
asking; asking after errors or contradicting repeats did not improve recovery.
The main study's larger asking benefit for kNN does not explain these failures.
A recent-window learner missed its primary criterion after changes on new contexts
($-0.010$ $[-0.046, +0.023]$) and slightly reduced fixed-rule performance.
Its gain with repeats ($+0.048$ $[+0.006, +0.090]$) was a secondary, exploratory result.
The staleness rule addresses memory; rule-level recovery remains unresolved.

\paragraph{Extensions.} Adding an 8B LLM rule level cost $0.039$ $[-0.095, +0.001]$
on new contexts and gave no gain out of grammar ($-0.031$ $[-0.090, +0.011]$).
A frontier-model version has not been run. For skill choice, the ladder passed an
archived-repair check but lost to NAPPING on new decision-level situations;
we make no broader skill-choice claim.

\paragraph{Excluded studies.} Three studies failed their validity checks: the initial
VLM baseline (79.4\% agreement with known verdicts against an 80\% threshold), the
initial ILP configuration, and the asymmetric-harm safety study, whose Popper
tie-breaking depended on the Python hash seed. The first two were corrected in
later comparisons, with ILP configured on separate tuning sites. None of the failed
studies supports a comparative claim. At tenfold harm cost, no learner beat
never acting on new scenes in the invalid safety study; this provides no validated
safety evidence.

% End sections/F_negative.tex

% Begin sections/I_continuing.tex
\section{Continuing-service frontier comparison: protocol and accounting}
\label{app:continuing}
% src: continuous_ladder_20261110/PROTOCOL_v1.md, RESULTS_v1.md,
% results_v1/{freeze,summary,audit,extended_audit,artifact_manifest}.json.

\subsection{Execution and evidence}
Each method/stream had one persistent physics/control worker, supervisor, deployment service and learner
for all 24 jobs. Independent streams started separately. The R1 pelvis was fixed; inverse kinematics
and PD control moved its arms. Every job included a logged external object setup preserving robot
joints, controller targets, worker identity and simulation time. Continuity applies within a stream,
not to autonomous replenishment or process-restart recovery.

Both providers used colour/shape segmentation and depth backprojection from $640\times400$ RGB-D
frames with public calibration. A frozen SmolVLA encoder supplied a 1,984-dimensional
vision/connector/depth vector for memory. Standardisation used 16 disjoint development frames;
the R1 domain did not reuse the main confirmatory study's image statistics. The frontier consumed perceived facts and
retrieved examples, not raw RGB. Paired streams shared nominal assignments, but images could differ
because each robot retained its own history.

After deliberation, a fresh frame had to agree on presence and perceived positions within 0.02\,m;
dispatch required a frame less than two seconds old and a native joint-speed stop check.
Physical completion required the chosen object's centre within $x\in[0.27,0.45]$\,m, within the table's
lateral bounds and above its surface, with other present objects displaced by at most 0.02\,m.
Private simulator state verified effects and evaluated the hidden requirement on the pre-action
scene; it was never a provider input. Acceptance also required the service's original-goal
verification. There was one physical attempt per job, with no retry.

\subsection{Providers and feedback}
The ladder preserved the executable semantics of the main confirmatory study: $\eta=0.002$, $\varepsilon=0.05$,
$k=3$, no recency decay, mixture prior $(0.6,0.2,0.2)$ and strict gate $d<0.01$.
The R1 grammar contained 271 hypotheses. A comparison against the original implementation matched predictions, weights and evidence
across 36 updates. The variant evaluated in the staleness study was not added.

The frontier was \texttt{gpt-6.1-sol}, medium reasoning, through the Codex CLI. Its persistent state
included seven nearest and four recent labelled examples, a complete ledger of its own events,
and editable rule notes limited to 3,000 characters. It received the same public grammar vocabulary, coordinates
and thresholds, and could explain rules outside that grammar. Each inference returned
schema-constrained probabilities and updated notes; file and shell tools were disabled. A trusted
input projection and transcript audit enforced the information boundary, without a claim of
adversarial process isolation.

There were at most 20 scored calls per stream, each with a 60-second deadline, and a per-stream
stop threshold of 750,000 reported input-plus-output tokens; an in-flight call could exceed it.
Model failures remained failures without retries or substitutes. The recorded hosted model alias
is not an immutable weight snapshot.

Both methods received four full-menu commissioning queries and at most three further full-menu
questions per scored stream. The common asking band was $0.25<\max_a p(a)<0.75$; otherwise a provider
executed the maximum if at least 0.5, or abstained. After a question it selected the first acceptable
item in catalogue order. A question updated learning once; subsequent execution did not duplicate
those labels. Ordinary F1 feedback labelled only its own physically completed action.

\subsection{Assignments and outcomes}
Three or four items were drawn from six coloured blocks/cups, with varied positions and an optional
white cue. All assignments had a nonempty acceptable set. Each stream contained four revisits;
other layouts were novel. Streams 3--5 changed requirement at zero-indexed job 14 without announcement.
Job 18 in streams 1 and 4 included a preassigned interruption after 0.15 simulation seconds if
an action was dispatched. Method order alternated by stream parity.

\begin{table}[htbp]
\centering\small
\caption{All six paired streams. Results are accepted / wrong dispatched / optional questions,
over all 20 scored assignments. Counts are descriptive.}
\label{tab:streams}
\begin{tabular}{clcc}
\toprule
Stream & Requirement & Ladder & Frontier \\
\midrule
0 & Leftmost & 18 / 0 / 0 & 18 / 0 / 0 \\
1 & Innermost & 17 / 1 / 0 & 15 / 2 / 2 \\
2 & Cue-conditioned left/right & 16 / 0 / 1 & 16 / 0 / 1 \\
3 & Rightmost $\to$ leftmost & 14 / 4 / 2 & 16 / 1 / 3 \\
4 & Cup-near-cue/fallback $\to$ cue inner/outer & 17 / 2 / 1 & 13 / 2 / 3 \\
5 & Second-left $\to$ not-innermost & 14 / 6 / 2 & 16 / 4 / 2 \\
\bottomrule
\end{tabular}
\end{table}

Stream 5's rules were outside the ladder grammar. Fixed-rule streams yielded 51/60 accepted jobs
for the ladder and 49/60 for the frontier; both accepted 22/24 revisited jobs. In stream 4, the
frontier's four fewer accepted jobs coincide with four model timeouts, with equal wrong-dispatch
and execution-failure counts. Some old and new requirements allow the same action on a scene;
post-change acceptance alone does not establish complete rule recovery.

\subsection{Interventions, latency and utility}
Including commissioning, accepted completion was 117/144 versus 115/144; each method accepted 21/24
commissioning jobs. Each received 24 commissioning queries supplying 84 labels and required 144
external setups. Optional queries supplied 22 versus 41 labels; F1 supplied 102 versus 92, giving
208 versus 217 total labels. Extra assistance and retries were zero by protocol, not because failed
tasks were repaired autonomously.

\begin{table}[htbp]
\centering\small
\caption{Declared utility sensitivity: accepted completions minus completed wrong actions minus
$c$ times optional questions. Setup, commissioning, F1 annotation, compute and delay are unpriced.}
\label{tab:cost}
\begin{tabular}{lrrrr}
\toprule
Question cost $c$ & 0 & 0.25 & 0.5 & 1 \\
\midrule
Ladder & 84 & 82.5 & 81 & 78 \\
Frontier & 86 & 83.25 & 80.5 & 75 \\
\bottomrule
\end{tabular}
\end{table}

The ranking crosses at $c=0.4$. Feedback counts are simulated oracle events, not human minutes or
money. Scored jobs took 11.02 minutes for the ladder and 71.93 for the frontier. Its 120 model calls
had mean latency 30.70 seconds (median 28.75); seven reached the deadline. Simulation targeted four
times real time and included CLI overhead, so these times do not estimate hardware throughput.

Audited reported usage was 2,532,054 input and 75,146 output tokens. Six timeout calls had no usage
receipt, so totals are incomplete. One other timeout had a receipt in partial output recovered by
the post-run parser; the frozen driver's accumulator omits it. Both totals are retained, and no
timed-out outcome was reclassified.

\subsection{Checks and diagnostics}
The prospective skill screen met its predeclared bar at 14/16, retaining both failed pushes.
Six contract checks preceded evaluation, including original-learner equivalence, future-information
isolation, failed-execution label suppression and exactly-once feedback. The final frontier prompt
passed disjoint induction and stale-note correction checks.

All 288 assignments finished in 87.62 minutes within the frozen 90-minute cap, without selective
reruns or omissions. Audits checked 281 physical receipts with finite monotonic native traces,
569 RGB-D frames, 288 feature vectors, persistent sessions and object-only setup changes. All four
injected interruptions confirmed a stop and withheld F1 feedback. Source, model, asset,
normalisation, protocol and raw-artifact hashes matched; no model tool calls were observed.

All 25 natural execution failures, including six commissioning failures, stopped short of the
clear-zone boundary. A shared wrong-action case arose at a 1\,mm tie boundary: perceived positions
put two candidates within tolerance, but evaluator positions put them just outside it. It remains
wrong for both methods. The memory gate was available on 14 scored ladder contexts and selected
the action on 12; only one post-change wrong action was a selected memory override. These
post-hoc diagnostics identify failure modes, not causal component benefits.

% End sections/I_continuing.tex

\clearpage
% Begin sections/J_c9.tex
\section{Initial closed-loop demonstration: failure case}
\label{app:c9}
% src: scope_closed_loop_c9_20261107/DEMO_RESULT.txt; perception_grounded_skills_v1_20261105/MILESTONE_RESULT_v1.txt.
\begin{figure}[htbp]
  \centering
  \includegraphics[width=0.9\linewidth]{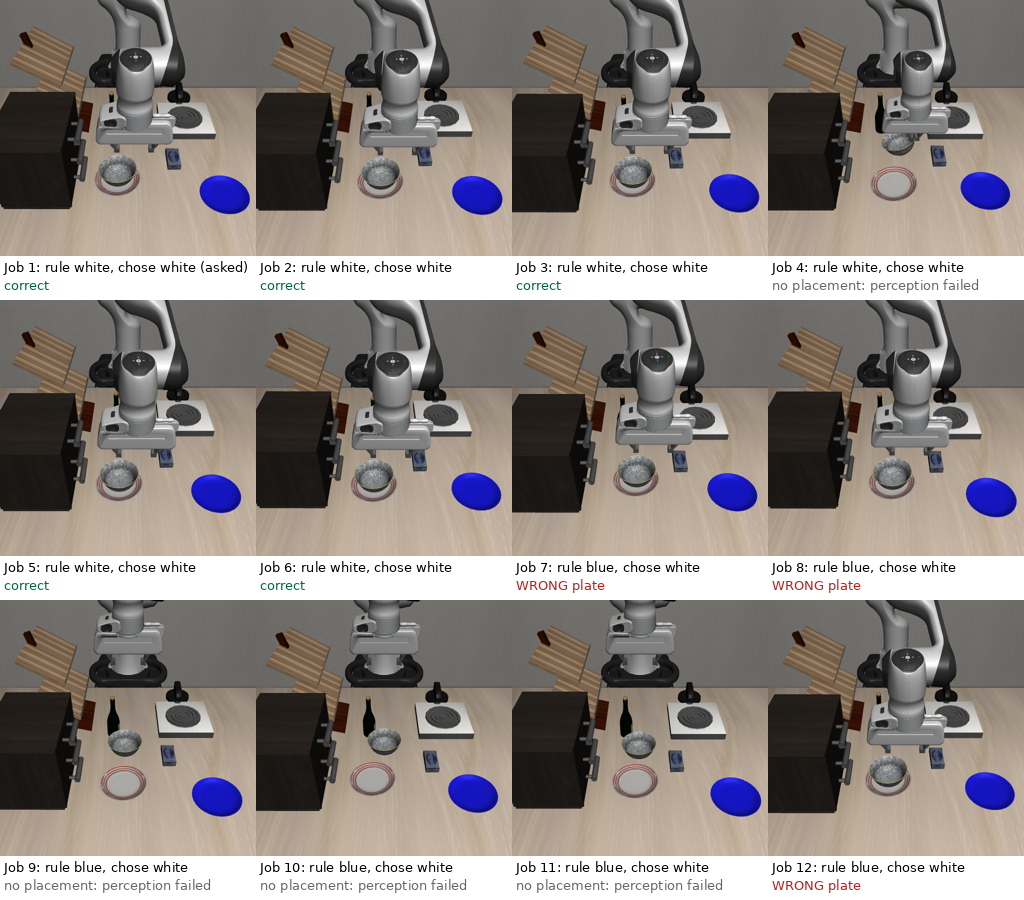}
  \caption{The initial closed-loop demonstration on the released skill layer: final frame of each of the 12 jobs. The
  hidden rule is ``white plate'' for jobs 1--6 and ``blue plate'' from job 7, without announcement. This single-stream
  demonstration is descriptive.}
  % src: scope_closed_loop_c9_20261107/demo_sheet.png
  \label{fig:c9}
\end{figure}
The initial closed-loop demonstration tested one 12-job stream on a frozen simulated Panda skill layer with five perception-grounded
skills, qualified on ten prospective episodes with eight passing independent checks
(\cref{app:skills}). The loop used the robot's camera, frozen probe/features, the main confirmatory study's ladder and F1 feedback;
each job launched a fresh simulator process. The hidden requirement was \emph{white plate} for six jobs
and then \emph{blue plate}, without announcement.

Starting without prior labels, the robot asked once on job 1 and chose the required plate on all
five physically completed pre-change jobs. After the change it kept choosing white: three wrong
placements and no further question, because confidence fell only to about 0.84, above the asking band.
This was a rule-level failure: all 12 perceived contexts were distinct, and the closest feature pair
had distance 0.097, above the gate threshold 0.01. The memory staleness rule would not have fired.
Three post-change jobs failed in perception and supplied no feedback.

The skill layer completed 8 of 13 physical attempts; all five failures were bowl-grounding
abstentions. This single-scene demonstration is a failure case, not a comparative reliability estimate.

% End sections/J_c9.tex

\clearpage
% Begin sections/K_droc.tex
\section{DROC plan-level adaptation: protocol and outcomes}
\label{app:droc}
% src: droc_comparator_20261110/{PROTOCOL_v1.md,METHOD_MAP_v1.md,adapter.py}.

\subsection{Published mechanism and adaptation boundary}
We used the released DROC implementation at commit
\texttt{33e3571c9efb} to specify the plan-level mechanism:
\texttt{replan} extracts task and robot constraints; \texttt{save\_plan\_info} appends numbered
entries; \texttt{retrieve\_plan\_info} selects relevant constraints semantically; the planner then
uses those constraints. Our independent implementation preserves that information flow.
The constraint bank has no CRAFT grammar, expert weights, near-instance gate or automatic
invalidation rule. New entries may explicitly record contradictions to earlier entries.

At a scored decision, one model call distils the pending part of the agent's own interaction
history and selects relevant indices from its old and newly appended constraints. A second
call receives the current camera-derived facts and only the selected constraints. It has no
access to the unselected bank or raw history. This contrasts with the custom frontier comparator, which
receives nearest/recent examples, a complete ledger of its own events and editable rule notes.
The pending history includes previous plans and factual feedback; no current verdict, hidden
rule, future context or other provider's observations are supplied.

The adaptation changes several parts of native DROC. Binary verdicts and explicitly budgeted
menu labels replace language corrections; the evaluator never supplies a reason for a verdict.
The motor controller is fixed, and each plan chooses one push rather than generating code or
replanning a multi-step task. Distillation and semantic selection share one model call, and
distillation is delayed until the next scored decision. The shared provider converts action
probabilities into a push, question or abstention. Known object identities remove the need for
DROC's optional ambiguous-object visual grounding; low-level DINO/CLIP retrieval and skill
corrections are not exercised. The model is \texttt{gpt-6.1-sol} at medium reasoning rather than
DROC's GPT-4. This \emph{DROC plan-level adaptation to verdict feedback} evaluates the published information flow
through our interface. It does not reproduce native DROC under its richer correction interface.

\subsection{Equal interaction contract and exposed controls}
The adaptation uses the unmodified service, provider, controller, camera frontend, normalisation
and evaluator from the continuing-service frontier comparison, on the same six nominal streams. Every stream starts a new robot and
learner that persist through its 24 jobs. The old ladder and frontier outcomes remain frozen
historical controls: they were not rerun concurrently or tuned for this follow-up. The streams
were already exposed, and hosted-model and native timing variation are possible. A new source
and assignment freeze precedes this implementation's measured outcomes; it does not turn the
exposed assignments into held-out evidence.

The four commissioning questions, three-optional-question limit, uncertainty band, one-attempt
physical budget, fresh-camera admission and failed-execution suppression are unchanged.
There are at most two model calls per scored job, sharing a single 60-second decision deadline;
neither stage has a reserved fraction. A failed call is not retried and the job remains a
failure. Successfully stored knowledge survives a later planning failure. Calls stop when the
reported stream token count exceeds 750,000 before a call; an in-flight call can exceed that
threshold. The whole run has a 150-minute cap, including execution and setup. The append-only
bank is not capped to match the frontier's 3,000-character editable notes; compute and memory
costs must therefore be reported alongside the shared interaction budget.

The original 14/16 physical qualification is reused because the motor and observation contracts
are unchanged. Six inherited and four additional CPU checks cover learner equivalence,
feedback suppression, public projection, selected-knowledge-only planning, append-only storage,
invalid retrieval indices and partial usage accounting. One disjoint, four-call model screen
passes spatial induction and contradiction handling before evaluation. No evaluation outcomes
are used to tune the prompts. All 144 assigned jobs, including 120 scored jobs, remain in the
denominators regardless of failure or non-start. Six streams are the descriptive replication
unit; no significance test against the historical controls is planned.

\subsection{Outcomes, costs and audit}
% src: droc_comparator_20261110/results_v1/{summary,audit,all_job_outcomes}.json(l).
% Begin sections/tab_droc.tex
% Generated by droc_comparator_20261110/posthoc/report.py from audited summaries.
\begin{table}[htbp]
\centering\small
\caption{DROC plan-level verdict adaptation, with frozen historical ladder and frontier controls from the continuing-service frontier comparison. All denominators are 120 scored assignments. This is an exposed-stream follow-up, not a fresh paired experiment or full DROC reproduction.}
\label{tab:droc}
\begin{tabular}{lrrr}
\toprule
Outcome & Ladder & Frontier & DROC-plan \\
\midrule
Accepted task completions & 96 & 94 & 93 \\
Physical completions & 108 & 102 & 104 \\
Wrong actions dispatched & 13 & 9 & 13 \\
Completed wrong actions & 12 & 8 & 11 \\
Execution failures & 12 & 11 & 14 \\
Injected failures (included above) & 2 & 2 & 2 \\
Provider errors & 0 & 7 & 0 \\
Optional full-menu questions & 6 & 11 & 12 \\
Abstentions & 0 & 0 & 2 \\
Unstarted assignments & 0 & 0 & 0 \\
\midrule
Utility, $c=0.5$ & 81 & 80.5 & 76 \\
\bottomrule
\end{tabular}
\end{table}

% End sections/tab_droc.tex

% Begin sections/tab_droc_streams.tex
\begin{table}[htbp]
\centering\small
\caption{DROC adaptation follow-up: per-stream outcomes, each over 20 scored assignments. Historical ladder and frontier stream results are retained in \cref{tab:streams}.}
\label{tab:droc-streams}
\begin{tabular}{crrrr}
\toprule
Stream & Accepted & Wrong dispatched & Questions & Provider errors \\
\midrule
0 & 19 & 0 & 1 & 0 \\
1 & 17 & 1 & 1 & 0 \\
2 & 16 & 0 & 1 & 0 \\
3 & 14 & 3 & 3 & 0 \\
4 & 15 & 3 & 3 & 0 \\
5 & 12 & 6 & 3 & 0 \\
\bottomrule
\end{tabular}
\end{table}

% End sections/tab_droc_streams.tex

All 144 assignments finished in 62.45 minutes, with no rerun, replacement, unstarted assignment
or unresolved service state. Accepted completion was 93/120 scored jobs and 114/144 including
commissioning. There were 14 scored execution failures, including two scheduled interruptions,
and two abstentions. All 15 natural physical failures across commissioning and scored jobs
stopped short of the clear zone. No model call failed or reached its deadline.

The 24 commissioning questions supplied 84 labels; 12 optional questions supplied 41, and
F1 supplied 93, for 218 labels in total. External setups numbered 144; extra assistance and
retries were zero by protocol. The 240 model calls reported 4,052,464 input and 60,900 output
tokens, with no missing usage receipts. Scored jobs took 60.05 minutes, including 49.26 minutes
of prediction. Final banks contained 17--38 constraints and 5,969--9,859 characters per stream.
These measured resource differences accompany the common interaction budget; they are not
estimates of human effort, monetary cost or hardware throughput.

The declared utility values at $c=0,0.25,0.5,1$ were $82,79,76,70$, respectively.
Fixed-requirement streams accepted 52/60 jobs, post-change jobs 19/30, and revisits 22/24.
Only two optional questions occurred after changes. The final stream contributed six wrong
dispatches and exhausted its question budget.
These slices describe where the configured adaptation struggled; they do not isolate a causal
component effect. In particular, a successful action after a change need not identify the new rule.
One wrong dispatch would satisfy the true rule evaluated on the perceived facts; it remains
wrong under the evaluator, rather than being excluded as a perception discrepancy.

The independent audit passed 6,516 checks over 2,820 raw-file hashes, 142 physical receipts,
286 RGB-D frames and 144 feature vectors. It recomputed effect/verdict scores, checked
exactly-once feedback and failure-label suppression, and verified persistent sessions,
monotonic native traces, source stability and selected-knowledge-only planner inputs.
Both injected interruptions stopped and withheld F1 labels. All model usage receipts and
errors are retained. Outside the declared menu queries, feedback labels only the agent's own
physically completed actions. No future contexts or other method's feedback entered the model inputs.

% End sections/K_droc.tex

\clearpage
% Begin sections/L_carryon.tex
\section{CarryOn planning-loop adaptation: fresh paired protocol and outcomes}
\label{app:carryon}
% src: carryon_comparator_20261111/{PROTOCOL_v1.md,METHOD_MAP_v1.md,adapter.py}.

\subsection{Native components and adaptation boundary}
CarryOn is a close system-level comparator because it connects persistent user and scene context
with execution around frozen policies \citep{wang2026carryon}. We use its released implementation
at commit \texttt{02c97b80987c}: the native \texttt{UserMemory} and \texttt{SceneMemory} stores,
\texttt{update\_user\_memory} and \texttt{update\_scene} skills, user/scene context blocks, and
\texttt{validate\_scene\_update\_requirement} are imported unchanged. User facts can be overwritten
by key or explicitly deleted; scene entries replace the last established content of a place.
The native user-summary prompt preserves its durability filter and instruction to retract
obsolete keys while retaining unaffected facts. No CRAFT expert weights, grammar inference,
near-instance gate or automatic contradiction invalidation are inserted.

At each scored decision, a summary call processes up to 40 pending events from the agent's own
interaction history and executes valid native user-memory writes or deletions. A separate planning call receives the
complete current user and scene memories, current camera-derived facts, and at most ten events from its own
dialogue history. Prior execution-status events are excluded from fresh-task planning, following
CarryOn's distinction between dialogue and past physical state. Literal labelled corrections
remain dialogue. The planner may also write or retract user facts and update scene memory before
returning action probabilities to the common decision layer. A newly observed or changed table
snapshot requires a scene write. The native validator rejects a plan that omits it; up to two
parser-repair calls share the decision's original deadline. A valid summary update survives a
later planning failure. Every operation and before/after memory snapshot is retained.

This remains a \emph{planning-loop adaptation}, with substantive differences from native CarryOn.
Sparse action verdicts and paid menu labels replace rich user dialogue; the evaluator supplies no
explanation, new rule or counterfactual alternative outside the declared query. The native
durability filter may omit ambiguous feedback from long-term memory; recent dialogue remains available.
One known visible tabletop and one push replace multi-step endpoint selection and hidden-object
search. The scene-update requirement is extended to changed visible-table snapshots, whose
content the planner authors. User summarisation is delayed until the next scored decision rather
than running asynchronously after task completion; its orchestration is reimplemented. Camera
facts replace raw image input, and \texttt{gpt-6.1-sol} at medium reasoning replaces the native
model backend. The full native agent loop, visual judge, shielding, recovery and skill reflection
are not executed. This experiment cannot establish performance against that complete system.

\subsection{Fresh pairing, common contract and prospective gates}
The fresh paired CarryOn comparison contains six new scene stream pairs, each with four commissioning and twenty scored jobs per
method: 288 assigned jobs and 120 scored per method. The scene generator uses a new frozen seed
range, and no generated scene matches those in the frontier comparison, its DROC follow-up or the
hand-authored validity fixtures. The six requirement cases themselves are inherited from the frontier comparison: fresh scenes are not independently authored
requirements or new rule families. Four planned revisits occur within each stream; three streams
change requirements without announcement. The same two scheduled execution interruptions are
assigned to each method. Pair order alternates between ladder-first and CarryOn-first streams.

Both methods start independent empty learners and persistent robot/service sessions. They use
the unmodified service, ladder, camera frontend, controller, normalisation, physical/task
verifiers and shared inquiry/dispatch rule from the continuing-service frontier comparison. The same four commissioning questions, three optional
questions, one physical attempt per job, failed-execution label suppression and fresh-camera
admission apply. The original physical qualification is reused because these contracts and
assets are unchanged. Object-only external setup is counted; robot state and simulation time
persist within streams. Historical frontier and DROC follow-up results are retained separately,
not substituted for fresh controls or pooled into this comparison.

Eight new and six inherited CPU checks cover native overwrite/deletion and retention, scene
replacement, the scene-write gate and repair, public projection, failed-execution label
suppression, summary persistence after planning failure, fresh-scene disjointness and parity with the main confirmatory study's ladder.
One disjoint model screen checks explicit preference acquisition/revision with unaffected-fact
retention, and spatial acquisition/revision from sparse verdict examples. All four decisions
pass in eight calls without tuning or a repeated screen. Protocol and source hashes are frozen
before measured observations.

CarryOn receives at most four calls per scored decision: summary, plan and up to two parser
repairs, all within a single 60-second deadline. Transport failures are not retried. The cap is
80 calls per stream and 480 overall. Calls stop when reported stream input/output tokens exceed
750,000 before a call; an in-flight call can cross the threshold. The global wall cap is 180
minutes. All failures, abstentions, unstarted assignments and incomplete usage receipts remain
in their respective denominators. Native memory size is not artificially capped. The primary
endpoint is accepted task completion; wrong dispatches, completed wrong actions, interventions,
execution failures, model cost and post-change/revisit slices accompany it. The comparison has six paired streams as its units and does not establish superiority or noninferiority.

\subsection{Outcomes and independent audit}
% Begin sections/tab_carryon_streams.tex
\begin{table}[ht]
\centering\small
\caption{Stream outcomes in the fresh paired CarryOn comparison: accepted completions / wrong dispatches / optional questions, each over 20 scored jobs. Streams 3--5 change requirements after job 13.}
\label{tab:carryon-streams}
\begin{tabular}{crr}
\toprule
Stream & Ladder & CarryOn-plan \\
\midrule
0 & 19 / 0 / 0 & 18 / 0 / 3 \\
1 & 19 / 0 / 0 & 18 / 1 / 2 \\
2 & 18 / 1 / 2 & 13 / 3 / 3 \\
3 & 15 / 3 / 1 & 16 / 1 / 3 \\
4 & 12 / 3 / 1 & 14 / 1 / 3 \\
5 & 13 / 5 / 3 & 11 / 3 / 3 \\
\bottomrule
\end{tabular}
\end{table}

% End sections/tab_carryon_streams.tex

% Begin sections/carryon_accounting.tex
% Generated from retained P3 summary and independent audit; no hand-entered outcomes.
The run recorded all 288 assignments in 91.92 minutes. Unstarted assignments numbered
0 for the ladder and 0 for CarryOn; unresolved jobs numbered
0 and 0. No measured assignment was rerun or replaced.
Including commissioning, accepted completion was 115/144 for the ladder
and 109/144 for CarryOn. The all-assignment denominators retain every
physical failure, abstention and model error.

Commissioning questions numbered 24 for the ladder and
24 for CarryOn, supplying 82 and
82 labels. Optional queries supplied 27
labels for the ladder and 61 for CarryOn; F1 supplied 100 and
81, respectively. Total labels were 209 and 224.
External setups numbered 144 and 144; extra assistance
numbered 0 and 0, and execution retries 0 and 0.
These are simulated interaction counts, not measured human effort.

CarryOn made 244 model calls, including 4 parser-repair calls. Reported usage
was 4,240,505 input and 97,963 output tokens;
9 calls lacked complete usage receipts. Missing usage is not imputed
as zero; token totals are reported usage, not measured monetary cost. Scored-job wall times were
10.68 and 76.58 minutes, including 0.01
and 66.60 minutes of prediction for ladder and CarryOn, respectively.
Native memory operations included 80 user-fact writes,
0 explicit retractions and 111 scene writes.
Overwriting an existing key is counted as a write, separately from explicit deletion.

Fixed-requirement streams accepted 56/60 and
49/60 jobs; post-change slices accepted
17/30 and 22/30;
revisits accepted 18/24 and
20/24. These are descriptive slices of the same six
pairs, not independent replications or proof of complete requirement identification.

The independent audit passed 11,895 checks over 3,937 raw-file hashes,
275 physical receipts, 563 RGB-D frames
and 288 feature vectors. It rescored physical effects and acceptability,
checked exactly-once feedback and failed-execution suppression, and verified persistent
native sessions and monotonic finite trajectories. It also replayed the actual upstream
memory operations from empty stores, reproducing each committed summary/plan memory state
and checking every model input against the agent's own prior feedback. Frozen local and
upstream source hashes remained unchanged.

% End sections/carryon_accounting.tex

% End sections/L_carryon.tex

\end{document}